\documentclass[11pt,fleqn]{article} 
\usepackage{titlesec}

\usepackage[T1]{fontenc}
\usepackage{listings}
\usepackage{graphicx}
\usepackage{placeins}
\usepackage{hyperref}
\usepackage{appendix}
\usepackage{pythonhighlight}
\usepackage{float}
\usepackage{matlab-prettifier}
\usepackage{subcaption}
\usepackage{algorithmic}
\usepackage{algorithm}

\usepackage[margin=0.9in]{geometry}

\title{Analysis of ROC for Edge Detectors}
\author{Kai Yi Ji}
\date{\today}

\begin{document} 
\maketitle
\begin{abstract}      
    This paper presents an evaluation of edge detectors using receiver operating characteristic (ROC) analysis on the BIPED dataset. Our study examines the benefits and drawbacks of applying this technique in Matlab. We observed that while ROC analysis is suitable for certain edge filters, but for filters such as Laplacian, Laplacian of Gaussian, and Canny, it presents challenges when accurately measuring their performance using ROC metrics. To address this issue, we introduce customization techniques to enhance the performance of these filters, enabling more accurate evaluation. Through our customization efforts, we achieved improved results, ultimately facilitating a comprehensive assessment of the edge detectors.
\end{abstract}
\section{Introduction}
\label{intro}

In recent years, edge detection has been a fundamental task in computer vision and image processing, with numerous techniques and algorithms proposed to address this challenge. Traditional edge detection methods, such as the Sobel, Canny, and Roberts operators \cite{sobel,canny,roberts}, have been widely used for their simplicity and effectiveness. Additionally, advanced techniques leveraging deep learning algorithms have emerged, demonstrating promising results in edge detection tasks.

Despite the abundance of edge detection methods, the evaluation and comparison of their performance remain crucial for understanding their strengths and limitations. In this paper, we focus on measuring the performance of the most commonly used edge detection kernels. To achieve this, we employ receiver operating characteristic (ROC) analysis, a widely accepted evaluation metric in various domains \cite{roc}. By utilizing ROC curves and the corresponding area under the curve (AUC), we can assess the effectiveness of different edge detection algorithms and identify their strengths and weaknesses.

The objective of this research is to provide a comprehensive evaluation of edge detection techniques using ROC and AUC metrics. By analyzing the performance of various edge detection kernels, we aim to enhance our understanding of their capabilities and identify the most suitable algorithms for specific applications. The findings of this study will contribute to the advancement of edge detection methodologies and assist researchers and practitioners in selecting appropriate techniques for their image analysis tasks.

\section{Related Works}

The field of edge detection has witnessed significant advancements in recent years, with a wide range of techniques proposed to address this fundamental task in computer vision and image processing. Among the well-known methods are the Roberts \cite{roberts}, Sobel \cite{sobel}, and Laplacian \cite{laplaciancite} operators, which have been widely used for their effectiveness in detecting edges in images.

The Roberts edge detection algorithm utilizes a simple 2x2 kernel to compute the image gradient and identify edges. However, this method can produce noisy edges and may not be suitable for all types of images due to its simplicity and limited noise handling capabilities.

In contrast, the Sobel edge detection algorithm employs a larger 3x3 kernel designed to be more robust to noise and yield more precise edges. It utilizes separate kernels for horizontal and vertical edges, combining them to obtain the final edge map based on the magnitude of the gradients.

To enhance the edge detection capabilities, first-order Gaussian filters \cite{firstorder} are commonly used in conjunction with gradient operators such as Sobel and Roberts. These filters leverage the first derivative of the Gaussian filter to achieve edge sharpening and enhance edge detection performance.

The Laplacian filter \cite{laplaciancite} computes the second derivative of the image, enabling the identification of zero crossings associated with object edges. However, the Laplacian filter is sensitive to noise, potentially leading to faulty edge detection results. To address this issue, the Laplacian of Gaussian \cite{log} (LoG) was introduced, which applies a Gaussian filter to smooth the image and mitigate noise before applying the Laplacian filter.

It is possible to combine first-order derivative filters with Laplacian to achieve better sharpening, which achieved in the paper \cite{sobellap}. Through this work, we are also going to use similar technique to get appropriate measurement.

In addition to these methods, the Canny algorithm \cite{canny} is a well-known and widely used edge detection technique. It incorporates multiple steps, including Gaussian filtering for noise reduction, gradient magnitude computation, non-maximum suppression to refine edges, and hysteresis thresholding to identify true weak edges.

ROC (Receiver Operating Characteristic) \cite{roc} analysis is a widely used technique for evaluating the performance of binary classification algorithms, including edge detectors. It provides a comprehensive assessment of the trade-off between true positive rate (TPR) and false positive rate (FPR) at various classification thresholds.

For our evaluation, we will utilize the BIPED dataset \cite{soria2020dense}, which consists of 250 manually labeled outdoor images with dimensions of 1280x720 pixels. This dataset serves as a benchmark for computer vision tasks, including edge detection. In this paper, we will leverage the BIPED dataset to analyze ROC curves and assess the performance of various edge detection algorithms in relation to ground truth for three selected images.

We will MATLAB \cite{MATLAB} in this paper to detect edges. MATLAB is a programming language used for various purposes, including numerical analysis, edge detection and simulation.

\section{Edge Detection}

In this section, we will conduct an experiment to detect edges using mentioned filters in \ref{intro}. To calculate the magnitude of filters, we will be using the absolute value of gradients. We used this function because it costs less computation than square root, as shown in Listing \ref{magnitude}. 

\begin{lstlisting}[style=Matlab-editor, label={magnitude}, caption={Magnitude function}]
    function m = magnitude(x,y)
        %m = sqrt(x.^2 + y.^2); % more costly
        m = abs(x) + abs(y);
    end
\end{lstlisting}

When applying Roberts \cite{roberts} or Sobel \cite{sobel} filters to an image, the resulting edges can often be noisy. To address this issue, a common approach is to employ thresholding. Thresholding allows us to identify pixels with intensities stronger than a specified threshold as strong edges, while considering the rest as non-edges. By utilizing this technique, we can effectively reduce noise and enhance the clarity of the image.

For the First-order Gaussian method \cite{firstorder}, we begin by creating a probability density function that represents the Gaussian distribution. We then employ matrix multiplication to generate a Gaussian filter. This filter's smoothing effect can be adjusted by modifying parameters such as the standard deviation and mask size, thereby enabling customizations.

In the case of the Laplacian \cite{laplaciancite} operator, we convolve the image with a Laplacian filter. After this step, we can utilize a built-in edge detection function that detects zero crossings. This helps identify areas of significant change in intensity, which often correspond to edges in the image. Listing \ref{zerocross} demonstrates this.

\begin{lstlisting}[style=Matlab-editor, label={zerocross}, caption={Zerocross function}]
    laplacian_zerocross = edge(laplacian_image,'zerocross');
\end{lstlisting}

To optimize efficiency, we can combine the Laplacian filter \cite{laplaciancite} and the Gaussian filter to create the Laplacian of Gaussian \cite{log} (LoG) filter, rather than applying them sequentially. By doing so, we reduce the number of convolutions required in the image processing pipeline, leading to improved computational performance. Additionally, we employed the same edge function for the Canny algorithm \cite{canny}. This particular function has the capability to automatically select appropriate high and low thresholds for hysteresis thresholding. Hysteresis thresholding is a crucial step in the Canny algorithm, as it helps determine the final edges based on the strength of the detected edges and their connectivity. This is illustrated in Listing \ref{edgecanny}.

\begin{lstlisting}[style=Matlab-editor, label={edgecanny}, caption={Built-in Canny function}]
    canny_image = edge(image,'Canny');
\end{lstlisting}

\begin{figure}[H]
  \centering
  \begin{subfigure}[t]{0.32\textwidth}
    \includegraphics[width=\linewidth, height=3cm]{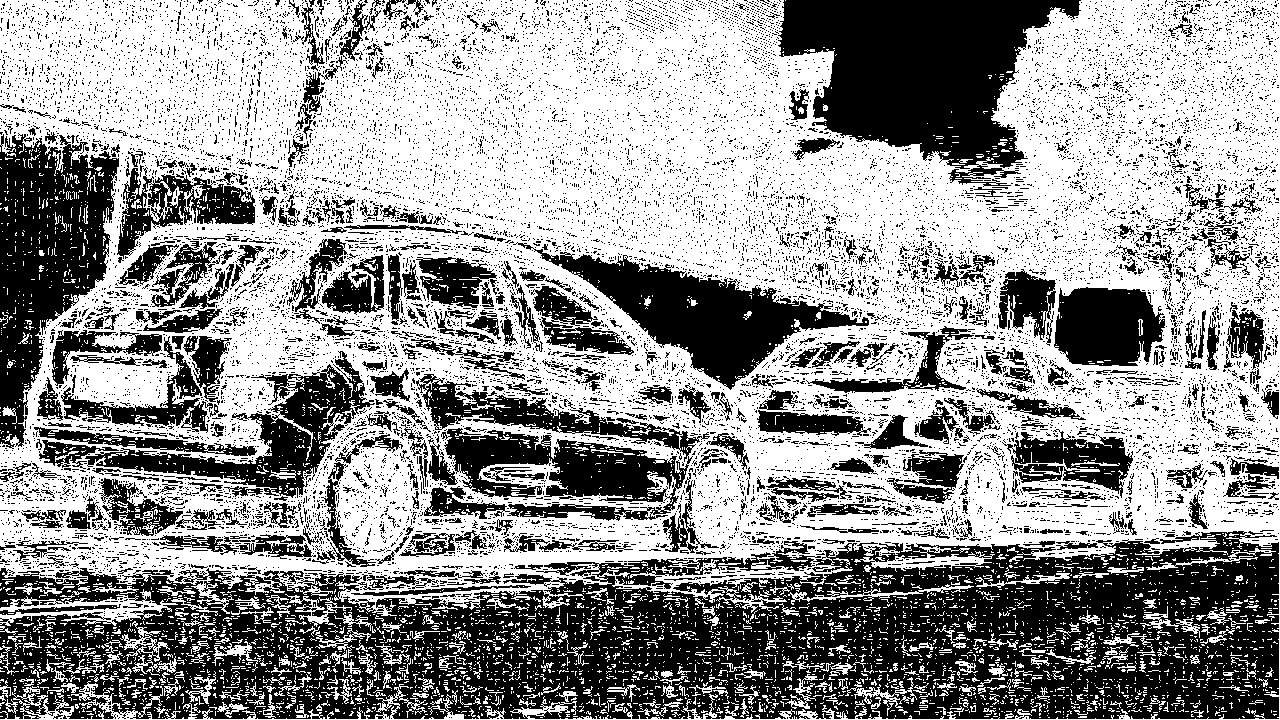}
    \caption{Roberts: RGB\_001}
  \end{subfigure}
  \hfill
  \begin{subfigure}[t]{0.32\textwidth}
    \includegraphics[width=\linewidth, height=3cm]{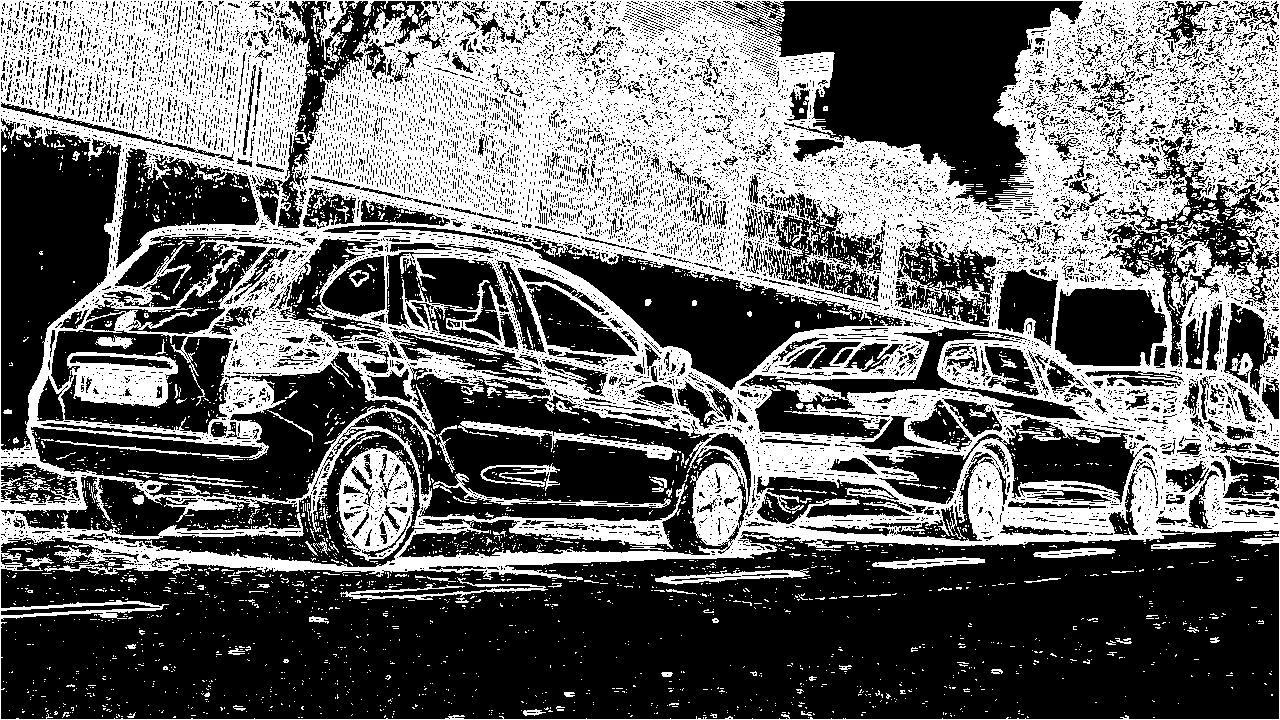}
    \caption{Sobel : RGB\_001}
  \end{subfigure}
  \hfill
  \begin{subfigure}[t]{0.32\textwidth}
    \includegraphics[width=\linewidth, height=3cm]{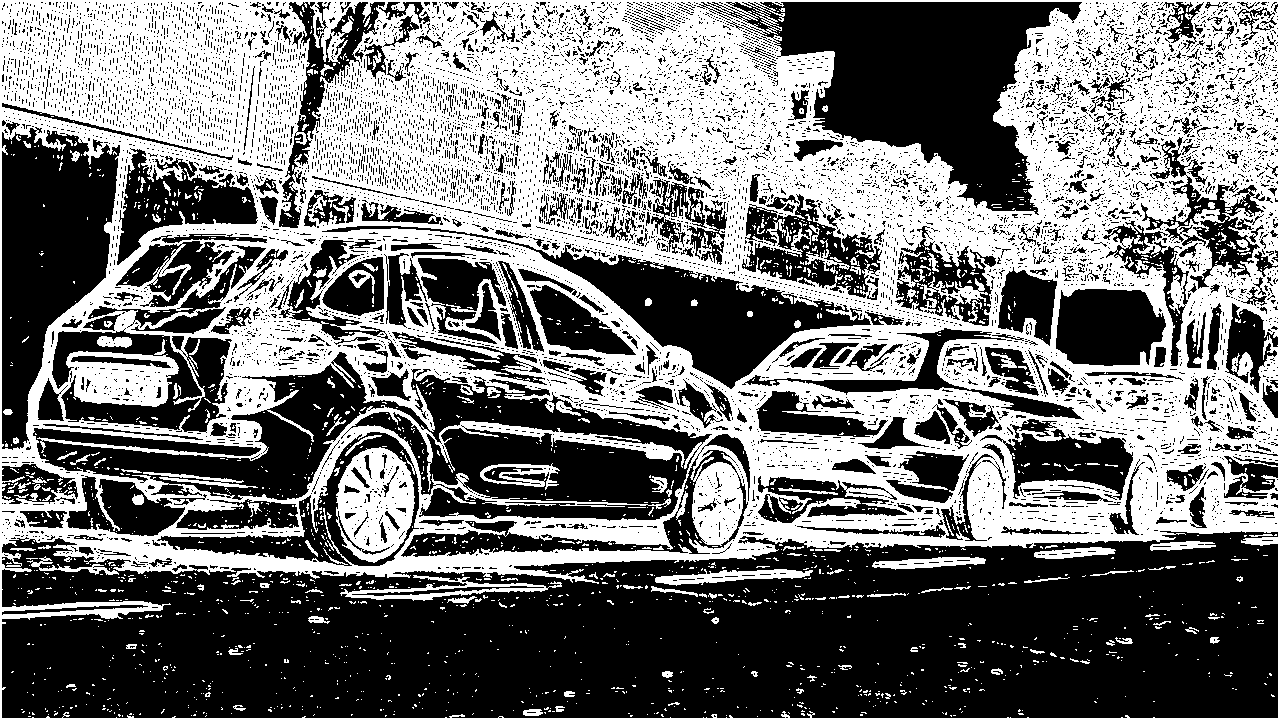}
    \caption{First-order Gaussian: RGB\_001}
  \end{subfigure}
  \begin{subfigure}[t]{0.32\textwidth}
    \includegraphics[width=\linewidth, height=3cm]{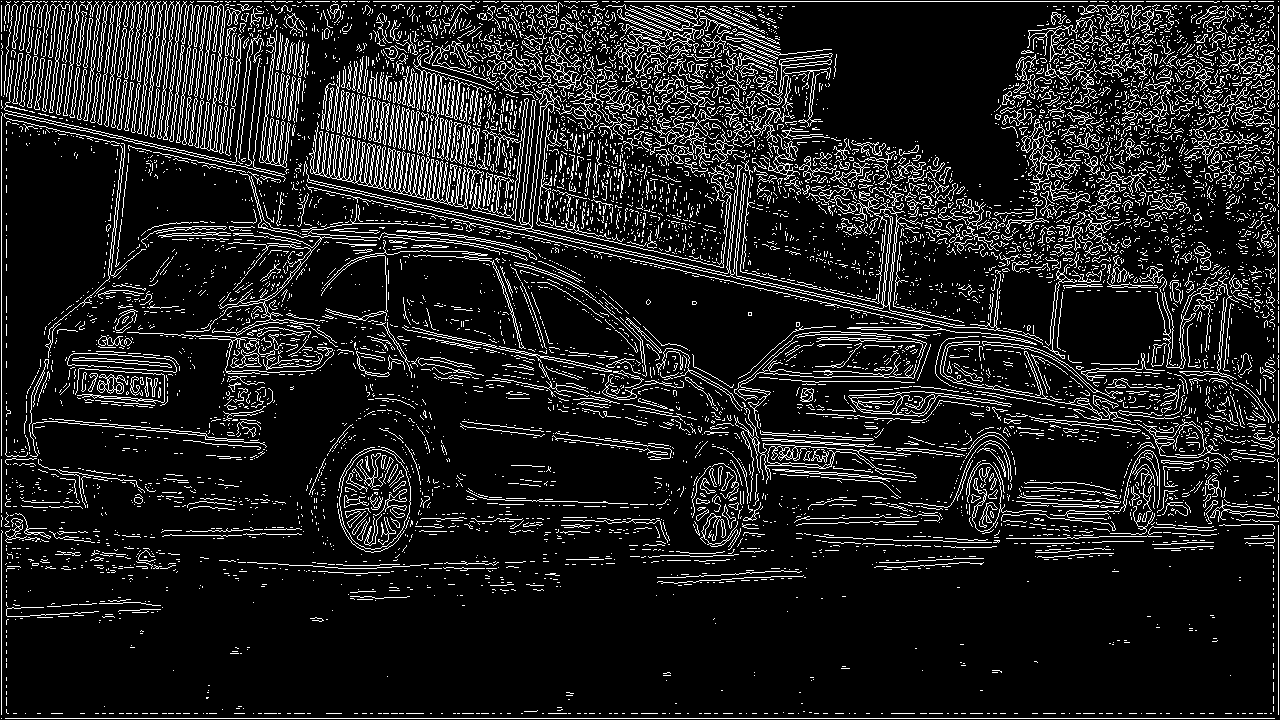}
    \caption{Laplacian: RGB\_001}
  \end{subfigure}
  \begin{subfigure}[t]{0.32\textwidth}
    \includegraphics[width=\linewidth, height=3cm]{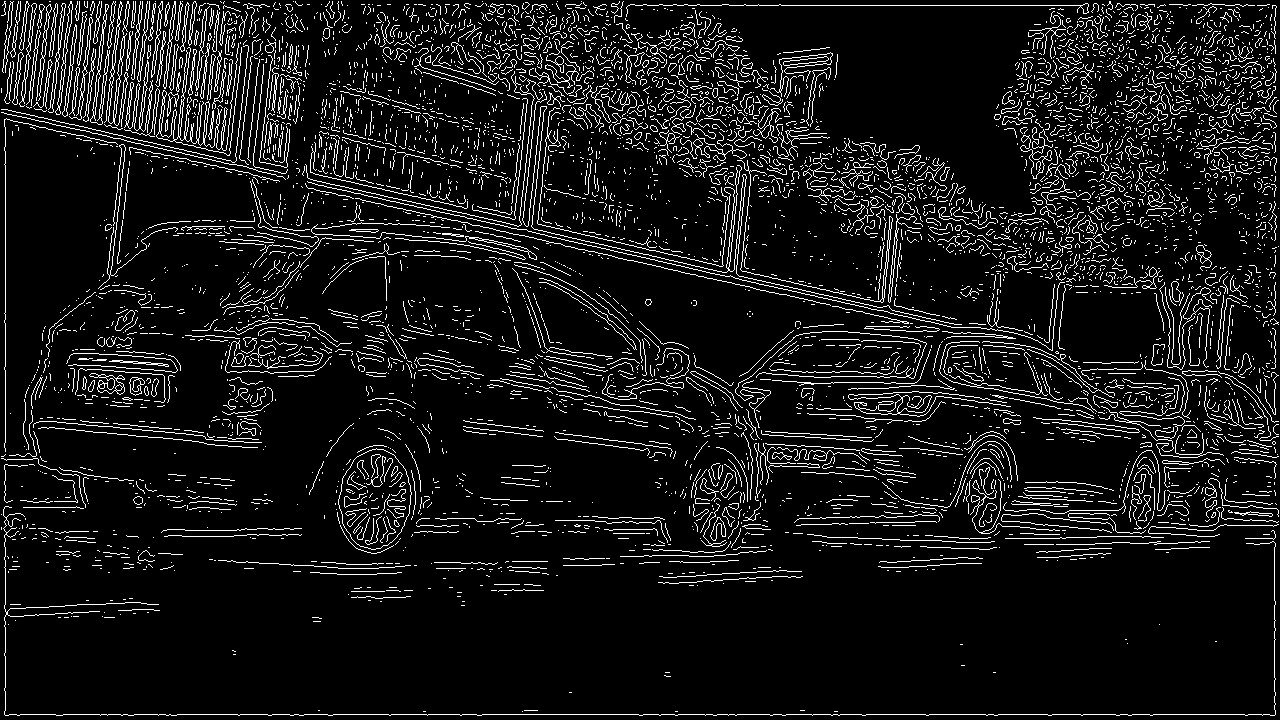}
    \caption{LoG: RGB\_001}
  \end{subfigure}
  \hfill
  \begin{subfigure}[t]{0.32\textwidth}
    \includegraphics[width=\linewidth, height=3cm]{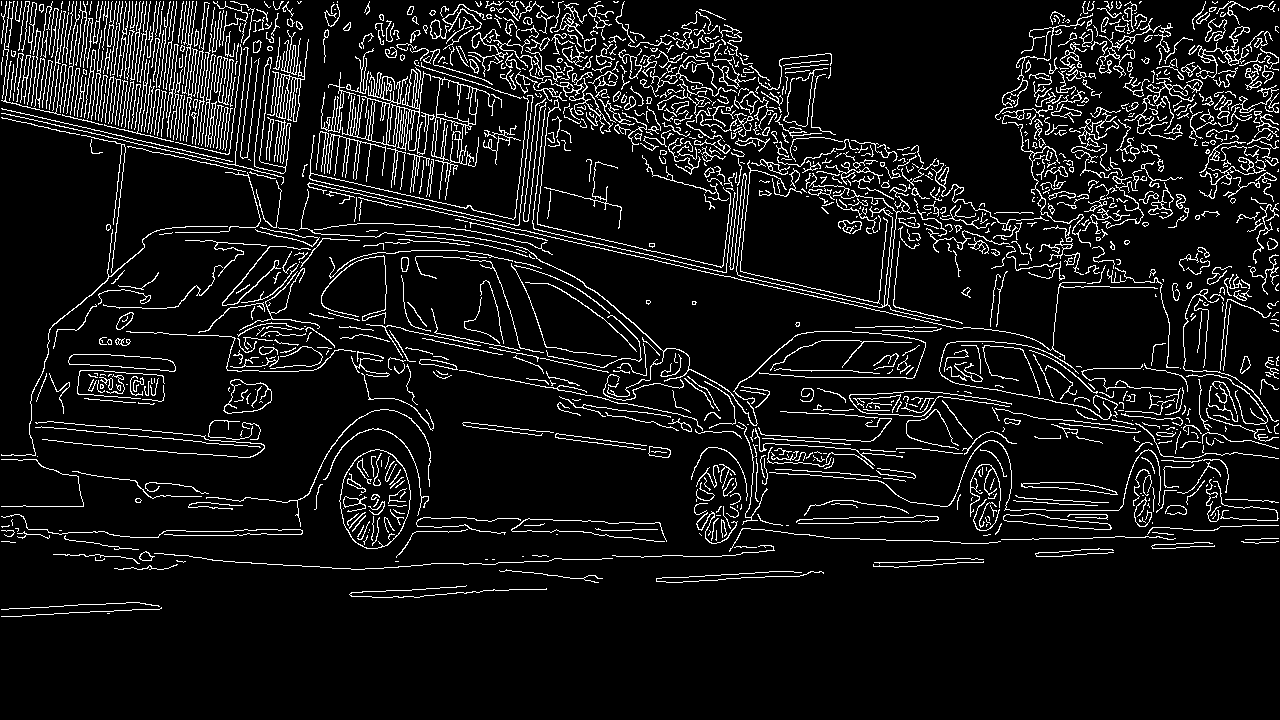}
    \caption{Canny : RGB\_001}
  \end{subfigure}

    \centering
  \begin{subfigure}[t]{0.32\textwidth}
    \includegraphics[width=\linewidth, height=3cm]{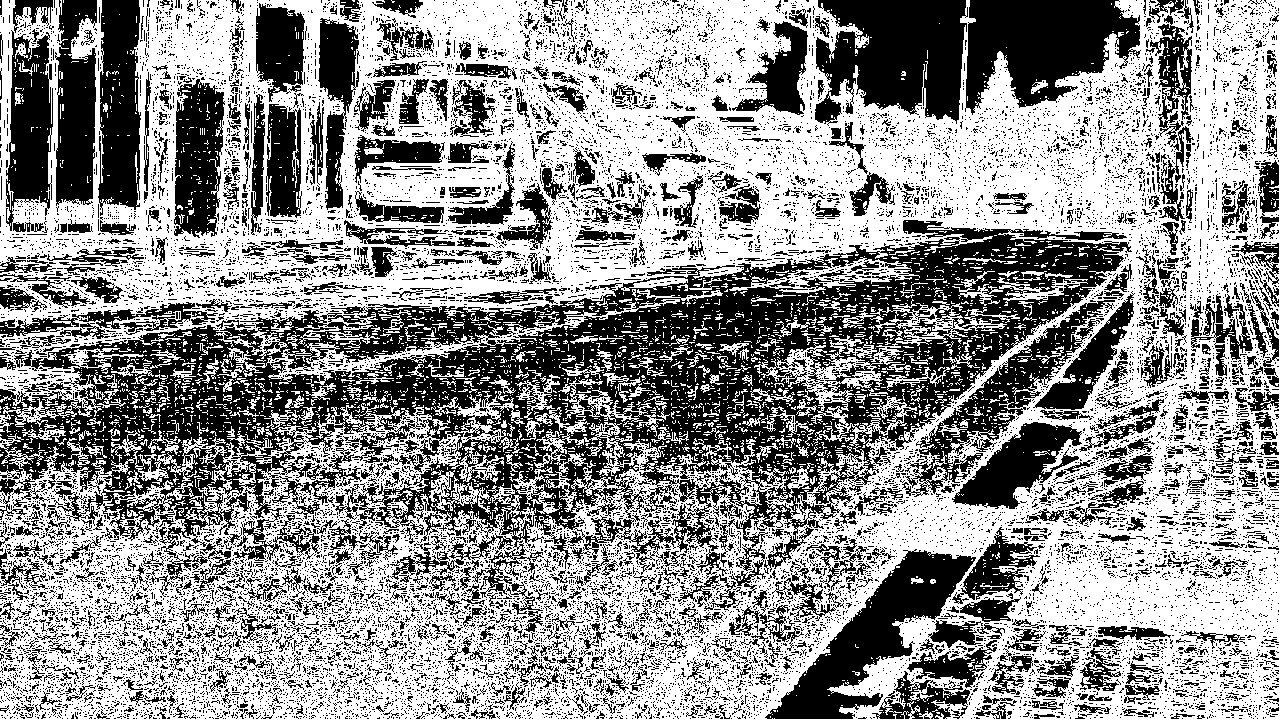}
    \caption{Roberts: RGB\_002}
  \end{subfigure}
  \hfill
  \begin{subfigure}[t]{0.32\textwidth}
    \includegraphics[width=\linewidth, height=3cm]{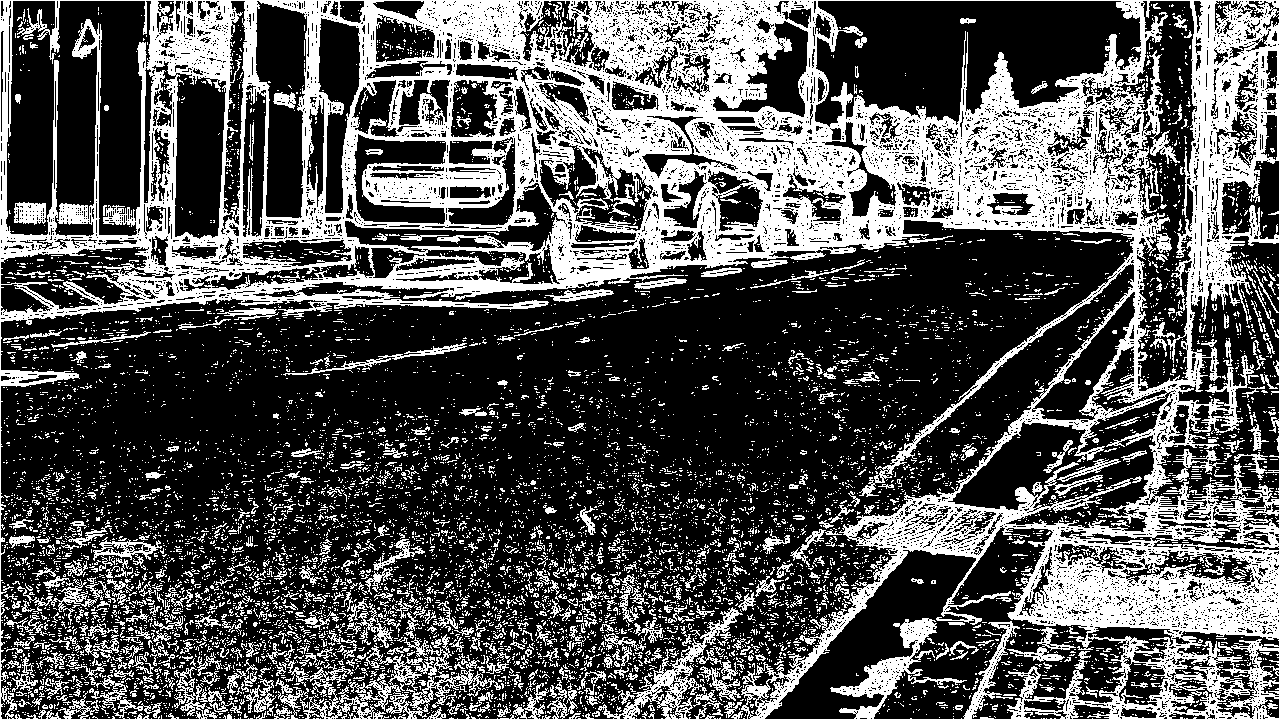}
    \caption{Sobel: RGB\_002}
  \end{subfigure}
  \hfill
  \begin{subfigure}[t]{0.32\textwidth}
    \includegraphics[width=\linewidth, height=3cm]{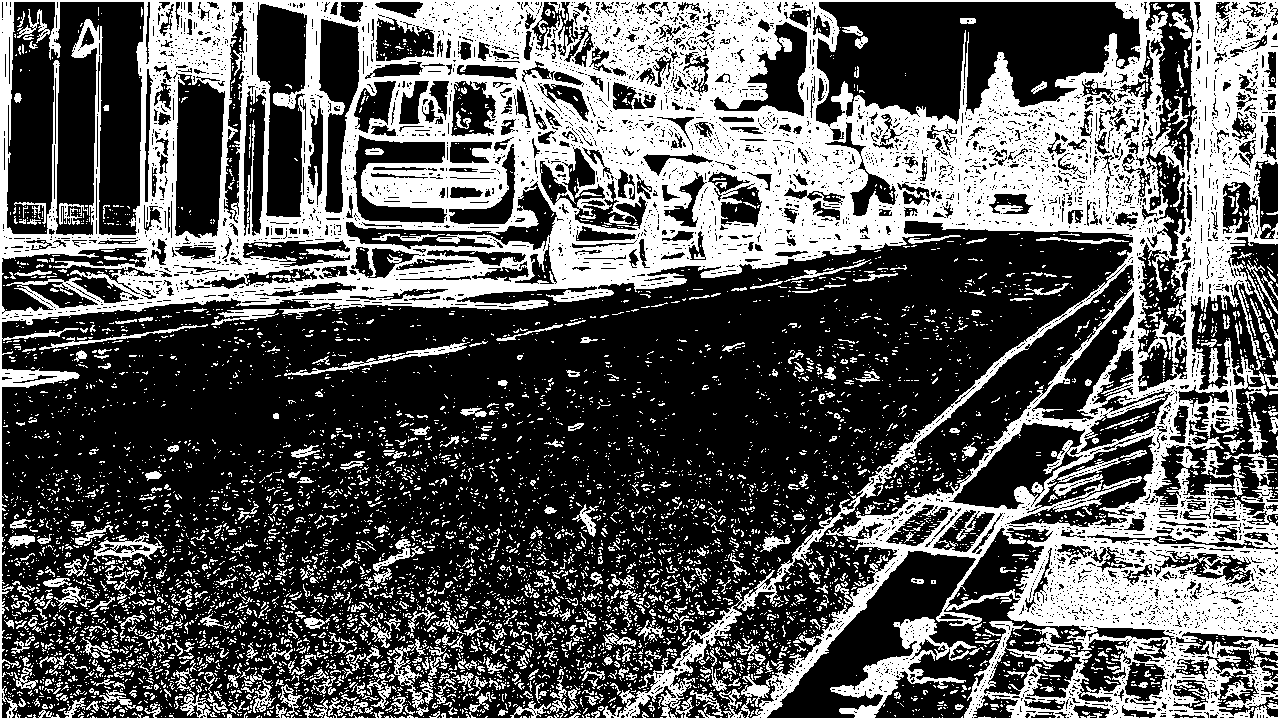}
    \caption{First-order Gaussian: RGB\_002}
  \end{subfigure}
  \begin{subfigure}[t]{0.32\textwidth}
    \includegraphics[width=\linewidth, height=3cm]{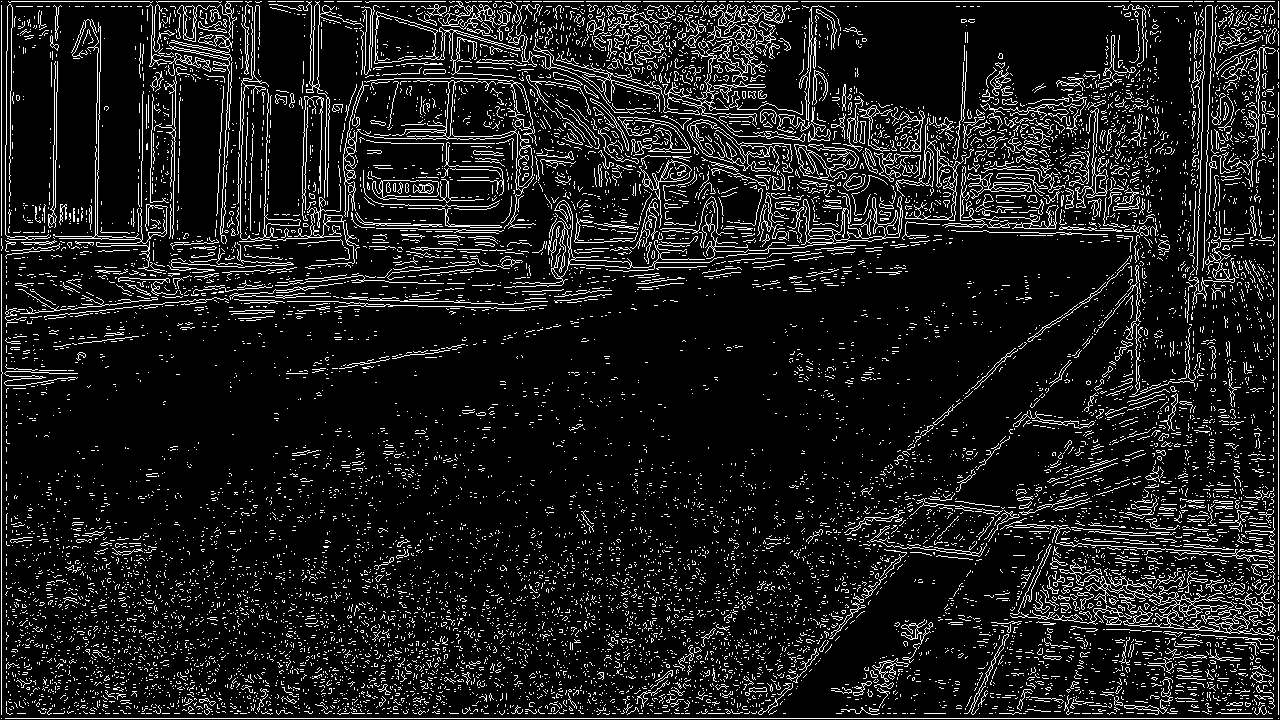}
    \caption{Laplacian: RGB\_002}
  \end{subfigure}
  \begin{subfigure}[t]{0.32\textwidth}
    \includegraphics[width=\linewidth, height=3cm]{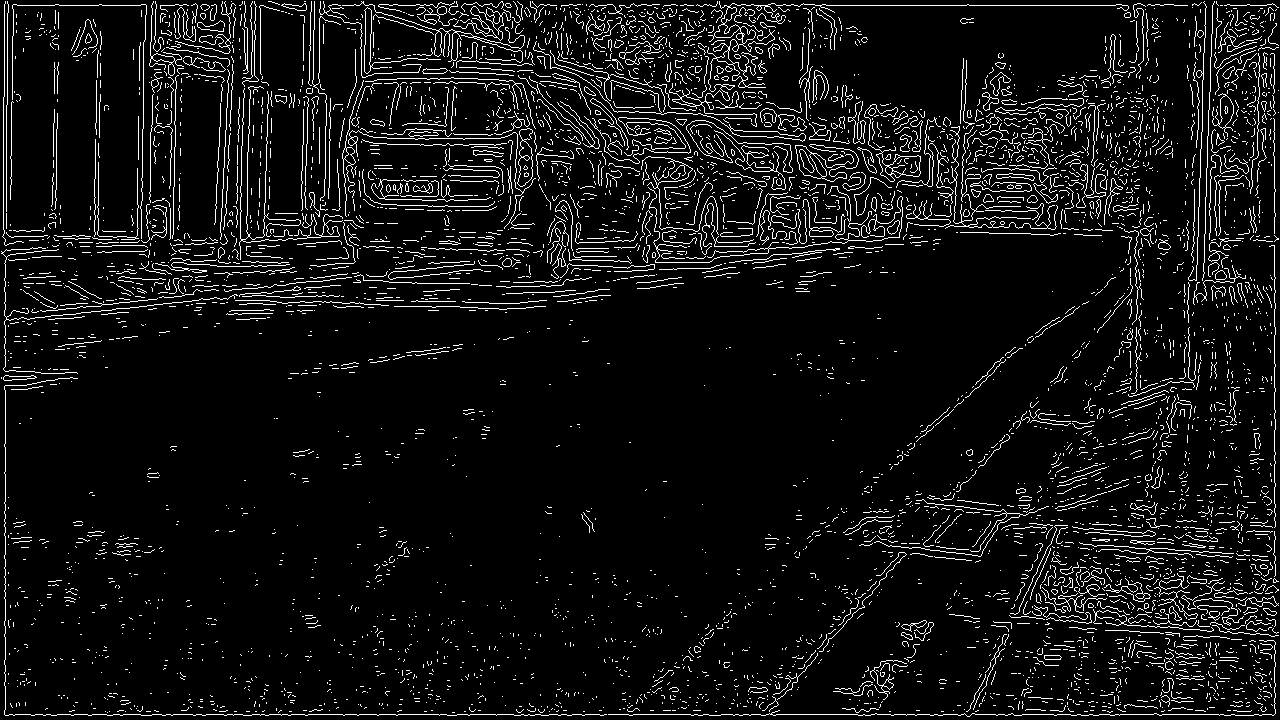}
    \caption{LoG: RGB\_002}
  \end{subfigure}
  \hfill
  \begin{subfigure}[t]{0.32\textwidth}
    \includegraphics[width=\linewidth, height=3cm]{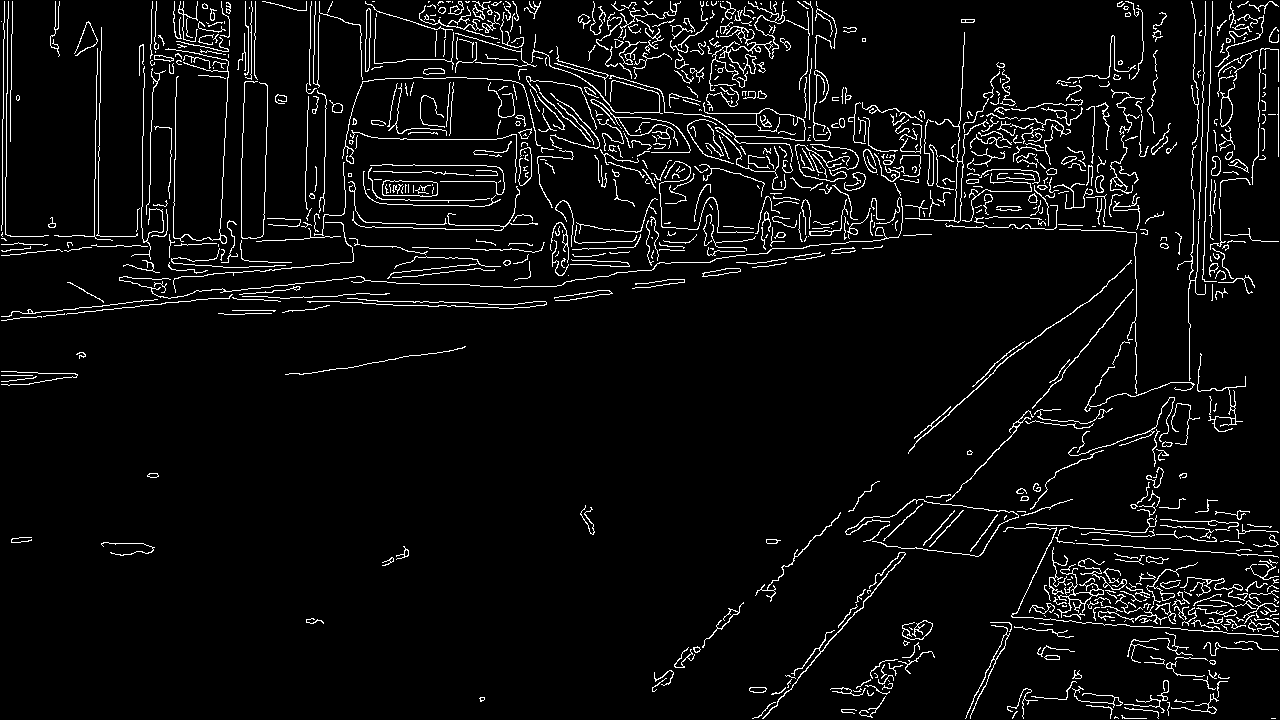}
    \caption{Canny: RGB\_002}
  \end{subfigure}
  \centering
  \begin{subfigure}[t]{0.32\textwidth}
    \includegraphics[width=\linewidth, height=3cm]{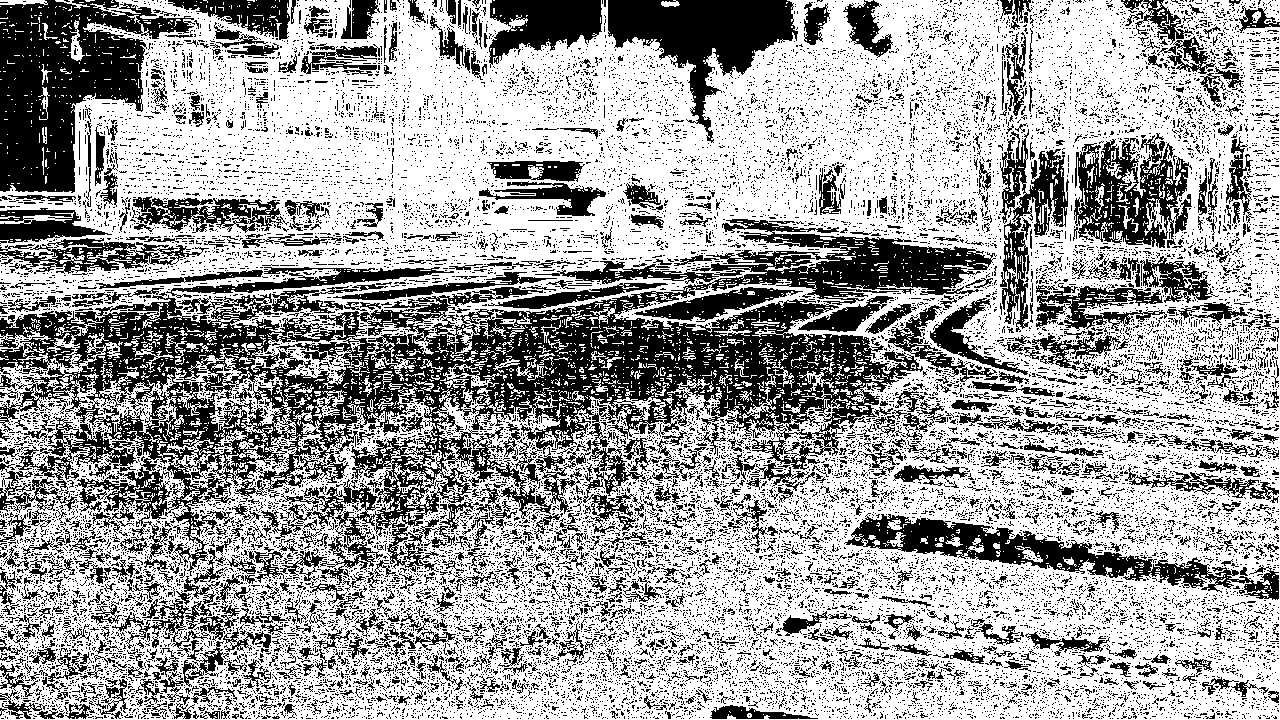}
    \caption{Roberts: RGB\_003}
  \end{subfigure}
  \hfill
  \begin{subfigure}[t]{0.32\textwidth}
    \includegraphics[width=\linewidth, height=3cm]{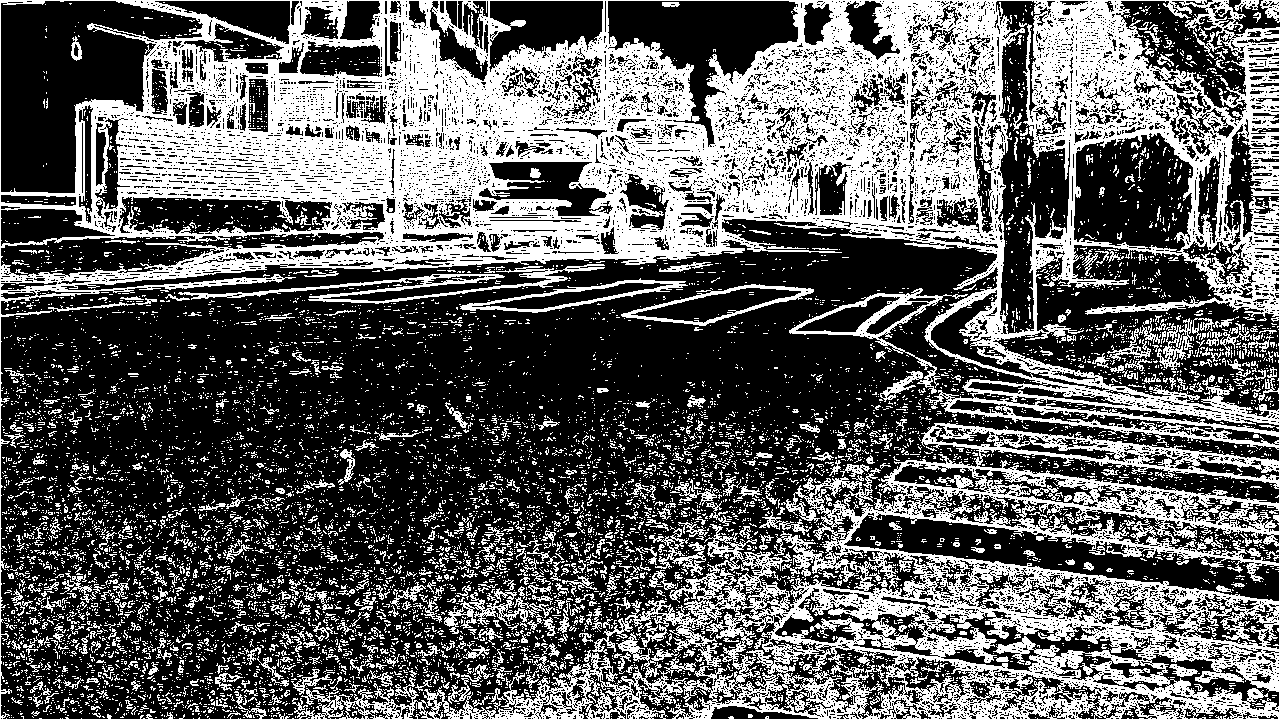}
    \caption{Sobel: RGB\_003}
  \end{subfigure}
  \hfill
  \begin{subfigure}[t]{0.32\textwidth}
    \includegraphics[width=\linewidth, height=3cm]{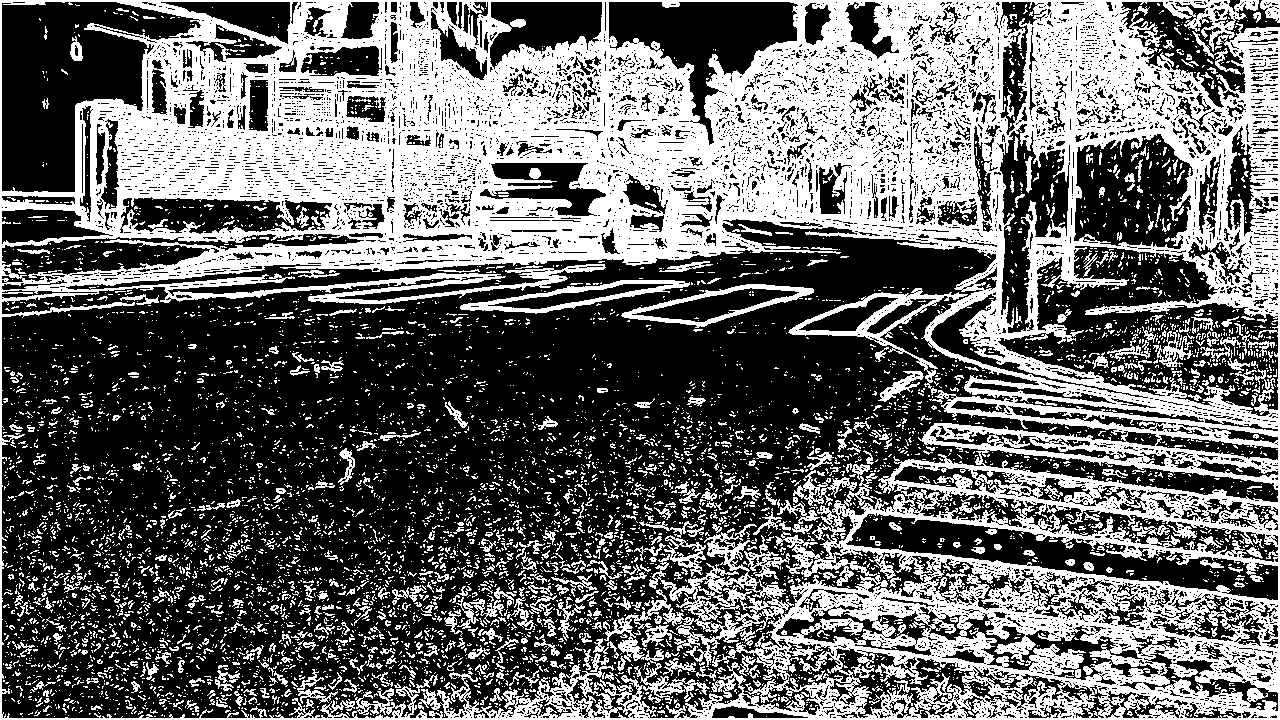}
    \caption{First-order Gaussian: RGB\_003}
  \end{subfigure}
  \begin{subfigure}[t]{0.32\textwidth}
    \includegraphics[width=\linewidth, height=3cm]{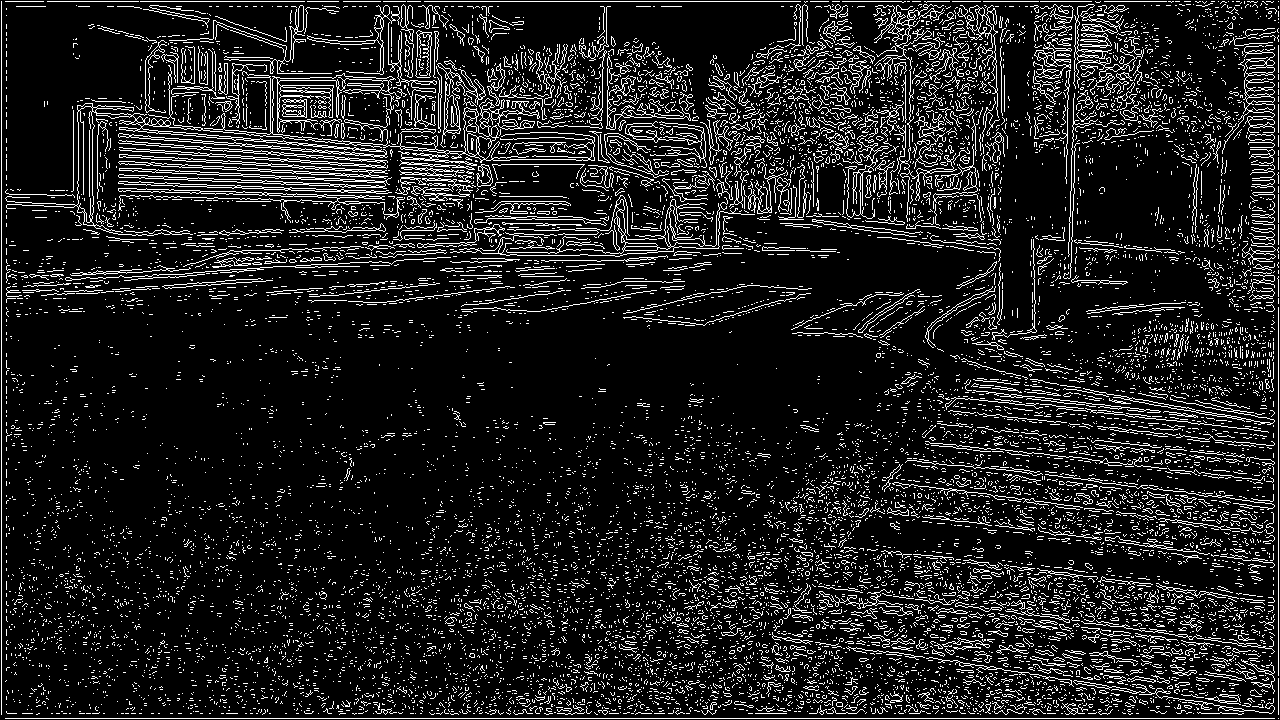}
    \caption{Laplacian: RGB\_003}
  \end{subfigure}
  \begin{subfigure}[t]{0.32\textwidth}
    \includegraphics[width=\linewidth, height=3cm]{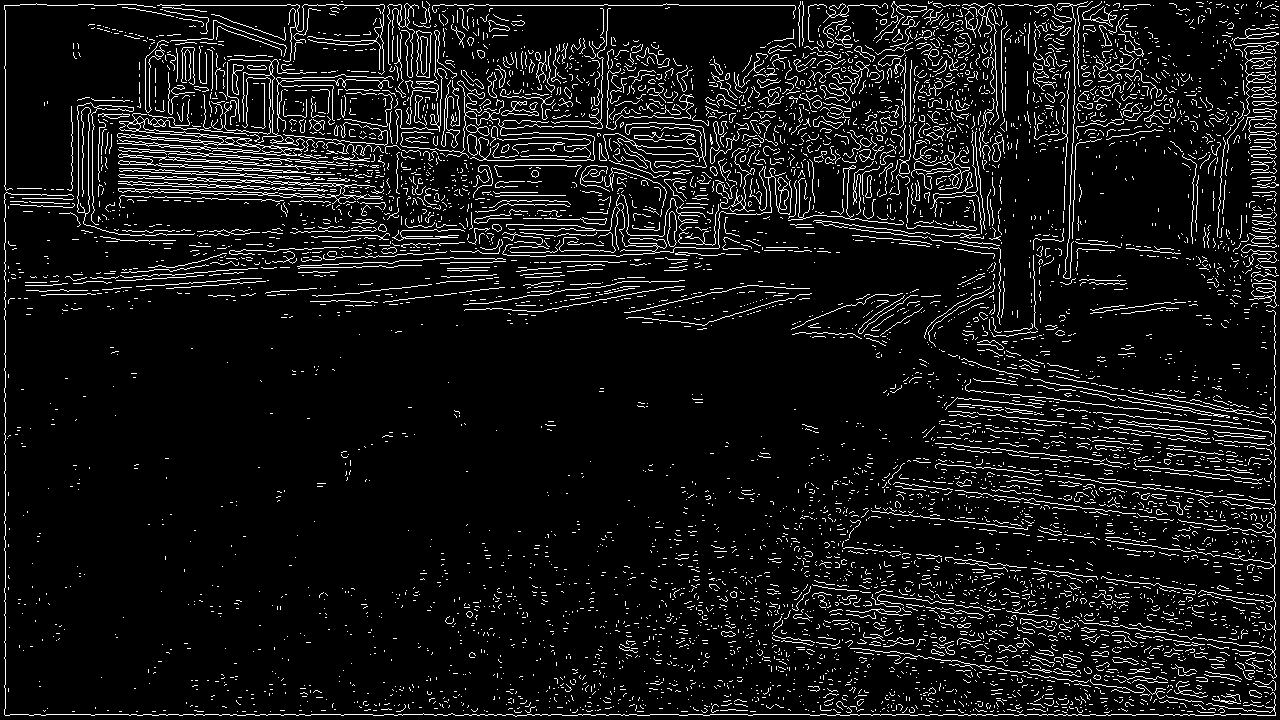}
    \caption{LoG: RGB\_003}
  \end{subfigure}
  \hfill
  \begin{subfigure}[t]{0.32\textwidth}
    \includegraphics[width=\linewidth, height=3cm]{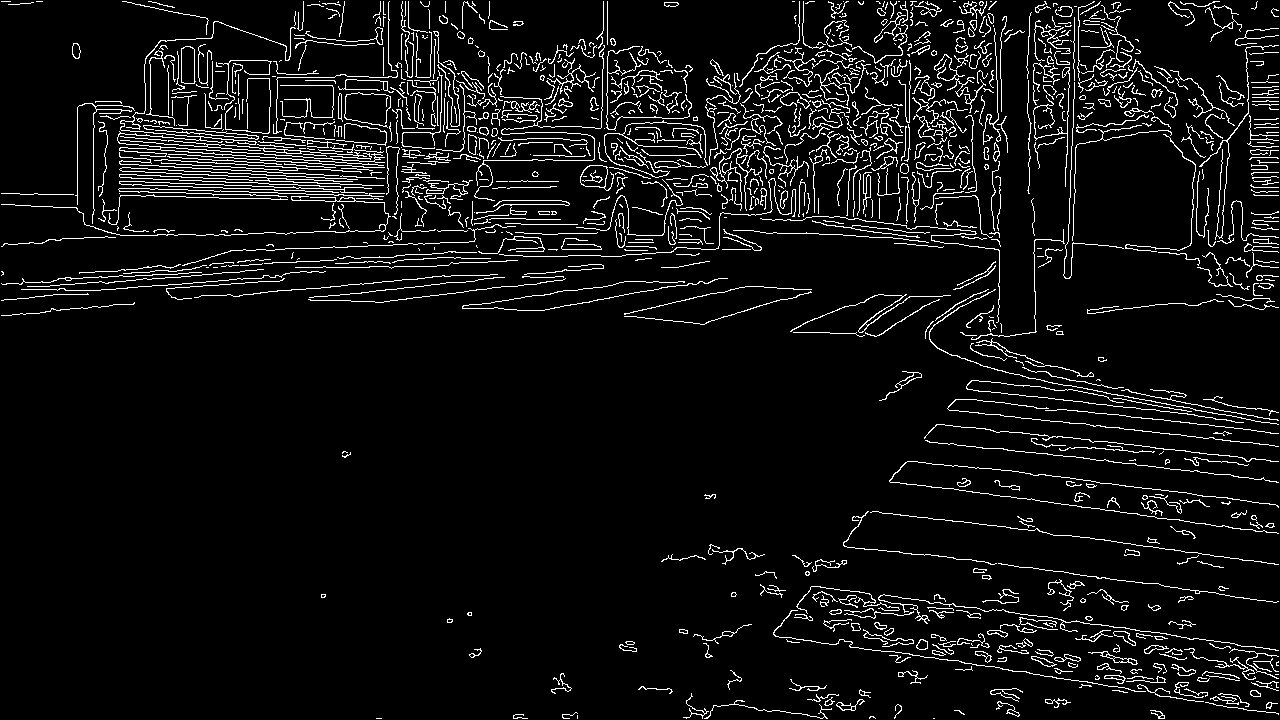}
    \caption{Canny: RGB\_003}
  \end{subfigure}
  \caption{Detected edges of all filters}
  \label{alledges}
\end{figure}

The results of all filters are shown in Figure \ref{alledges} for RGB\_001, RGB\_002 and RGB\_003. We observe that Roberts, Sobel and First-order Gaussian \cite{roberts,sobel,firstorder} had very noisy images. Roberts was noisiest and Sobel had less noise. First-order Gaussian produced a less noisy image but also had smoother edges. This is the effect of applying first derivative of Gaussian filter. Laplacian \cite{laplaciancite} filter was able to detect accurately but it is still prone to noise, which Laplacian of Gaussian \cite{log} was able to reduce. Lastly, Canny \cite{canny} was very accurate at predicting edges without much noise because it uses Gaussian filter to remove noise and then able to detect true weak edges using hysteresis thresholding.

\section{Receiver Operating Characteristic (ROC)}
\subsection{Background}
One commonly used method for evaluating the performance of edge detection algorithms is the Receiver Operating Characteristic (ROC) curve \cite{roc}. The ROC curve plots the False Positive Rate (FPR) on the x-axis and the True Positive Rate (TPR) or sensitivity on the y-axis. The TPR, also known as sensitivity, is calculated using Equation \ref{tpr}.

\begin{equation}
  TPR = \frac{TP}{TP + FN}
  \label{tpr}
\end{equation}

A good model will try to avoid false negatives to achieve a good TPR score. TNR or specificity is the measure how model performs in true negatives. A good model attempts to avoid false positives to perform well. This is shown in Equation \ref{tnr}.

\begin{equation}
  TNR = \frac{TN}{TN + FP}
  \label{tnr}
\end{equation}

FPR is calculated as in Equation \ref{fpr}. Informally, it measures the performs in terms of misses in TNR.

\begin{equation}
  FPR = 1 - TNR
  \label{fpr}
\end{equation}

To generate an ROC curve \cite{roc}, the output of a model is thresholded at various values. In our study, we utilized the perfcurve function in MATLAB to accomplish this task. This function automatically applies different thresholds to the edge detector and computes the corresponding False Positive Rate (FPR) and True Positive Rate (TPR) values. The Area Under the Curve (AUC) \cite{roc} is a metric that quantifies the performance of the edge detector. A larger AUC indicates a better-performing model. A perfect model would have a square AUC covering the entire ROC plot. Conversely, if the curve follows a diagonal line from the bottom left to the top right, it suggests that the model performs no better than a random guess.

\subsection{Initial Results}
\begin{figure}[H]
  \centering
  \begin{subfigure}[t]{0.40\textwidth}
    \includegraphics[width=\linewidth, height=6cm]{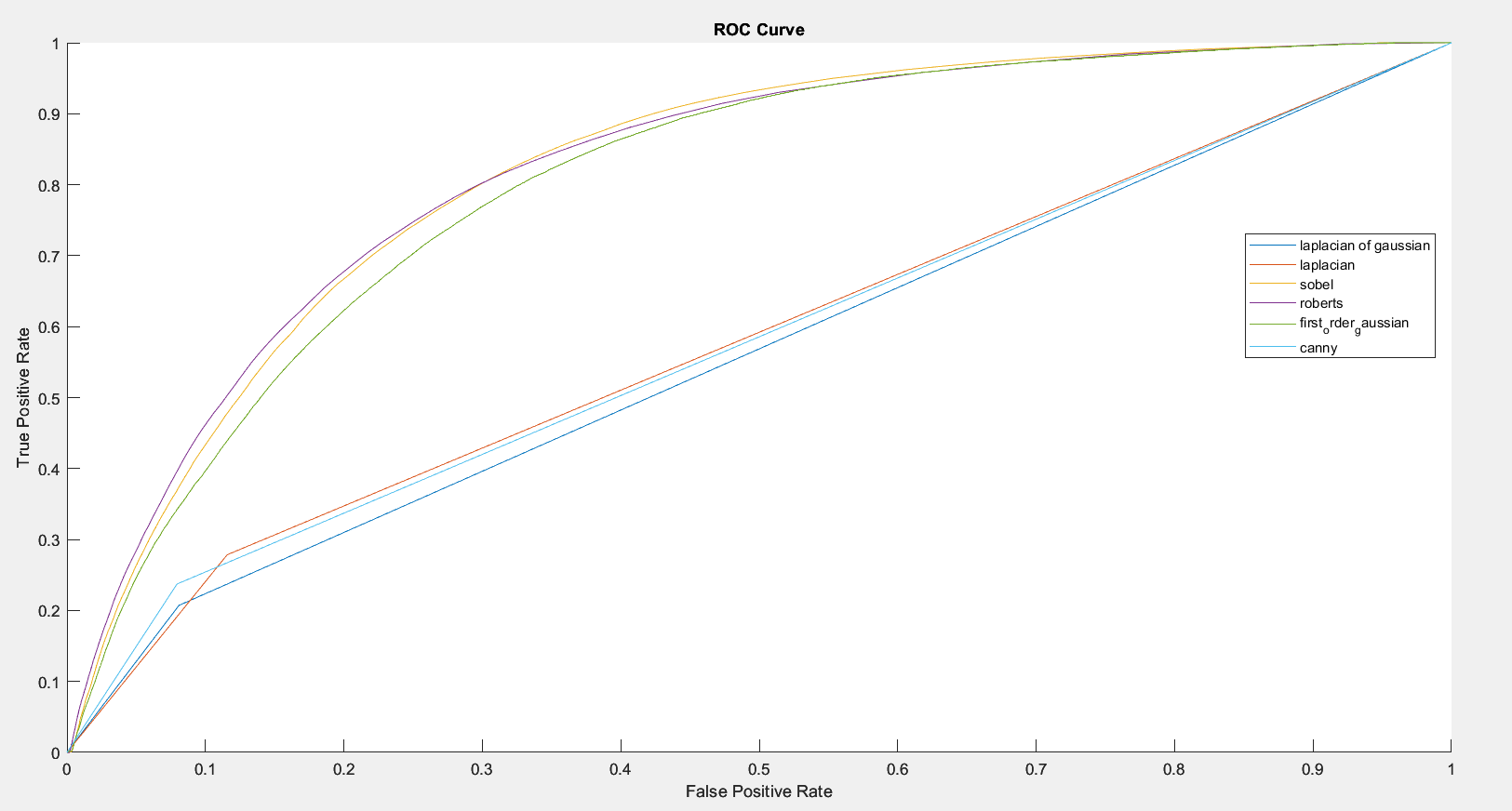}
    \caption{RGB\_001}
  \end{subfigure}
  \begin{subfigure}[t]{0.40\textwidth}
    \includegraphics[width=\linewidth, height=6cm]{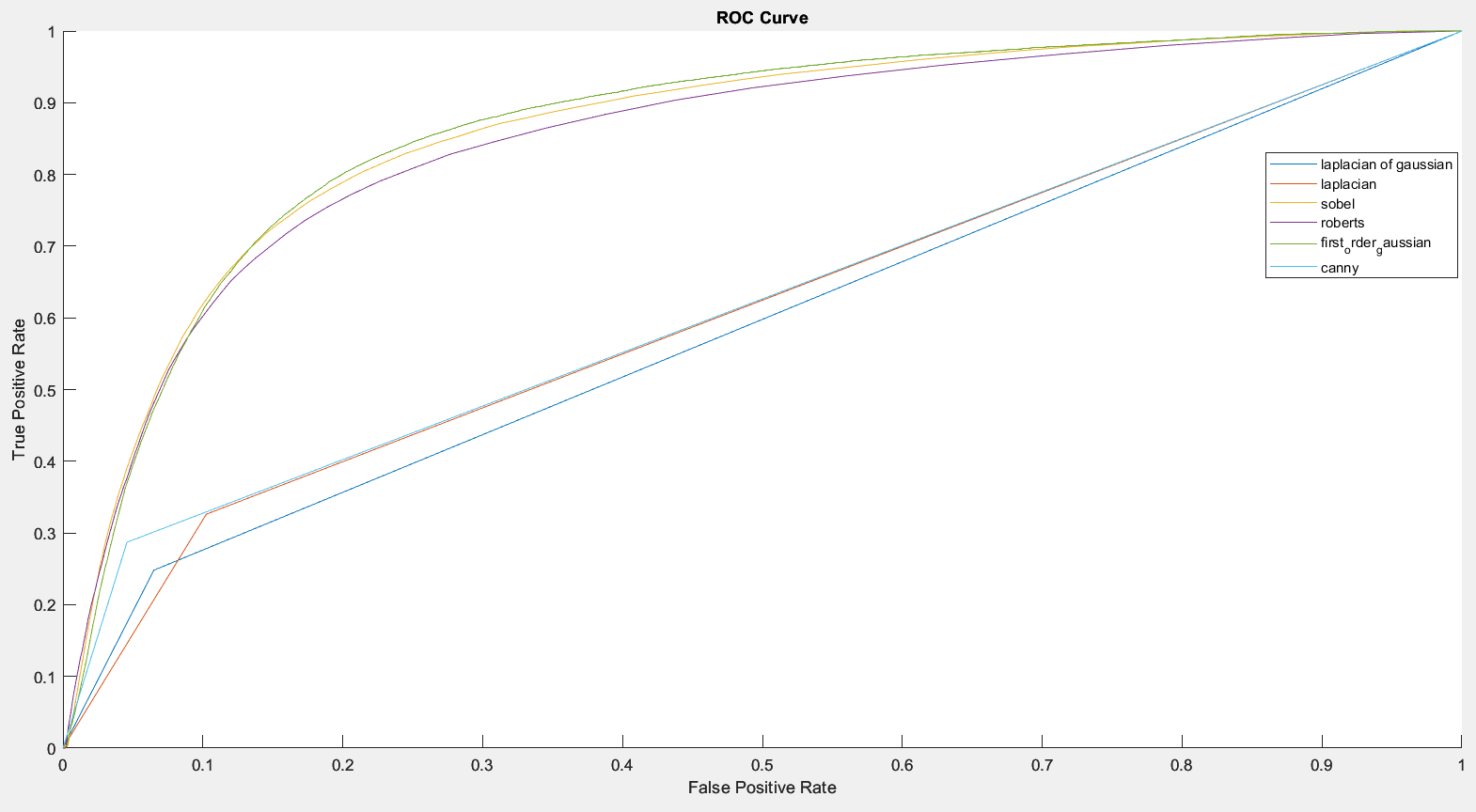}
    \caption{RGB\_002}
  \end{subfigure}
  \begin{subfigure}[t]{0.40\textwidth}
    \includegraphics[width=\linewidth, height=6cm]{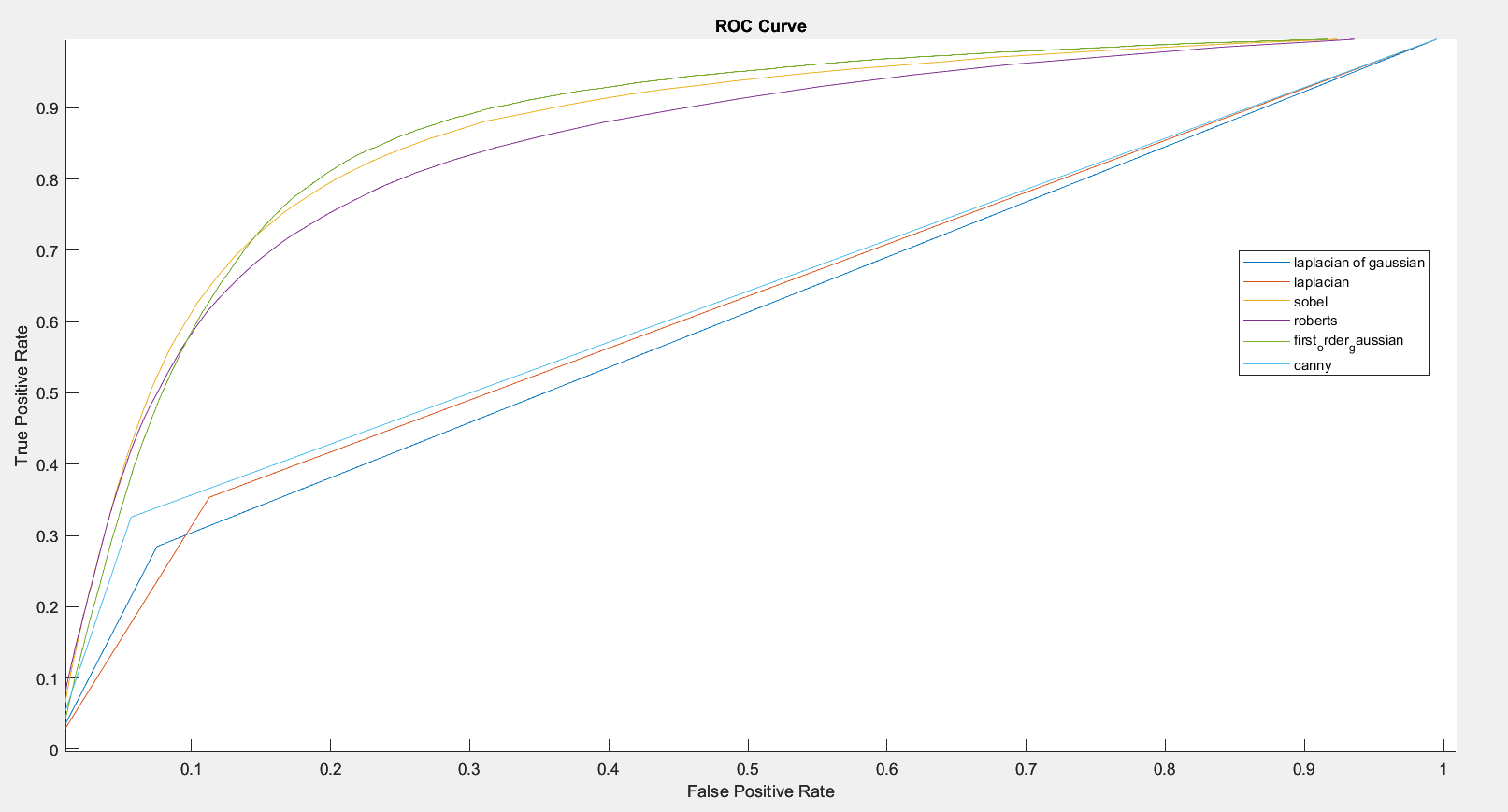}
    \caption{RGB\_003}
  \end{subfigure}
  \caption{Initial ROC Curve}
  \label{initroc}
\end{figure}

We plot the initial ROC curve and AUC for edge detectors in Figure \ref{initroc} and Table \ref{initauc}.

\begin{table}[H]
  \caption{Initial AUC of edge detectors}
  \begin{tabular}{ |p{2cm}||p{2cm}|p{2cm}|p{2cm}||p{2cm}|p{2cm}|p{2cm}|  }
      \hline
      \multicolumn{7}{|c|}{AUC} \\
      \hline
      Image & Roberts & Sobel & FoG & Laplacian & LoG & Canny \\
      \hline
      RGB\_001 & 0.82 & 0.82 & 0.80 & 0.58 & 0.56 & 0.58 \\
      RGB\_002 & 0.85 & 0.86 & 0.86 & 0.61 & 0.59 & 0.62 \\
      RGB\_003 & 0.84 & 0.86 & 0.87 & 0.62 & 0.60 & 0.63\\
      \hline
  \end{tabular}
  \label{initauc}
\end{table}

We see that for Laplacian \cite{laplaciancite}, Laplacian of Gaussian \cite{log} and Canny the curve is jagged. This is due to the edge function returns binary output. Because of this on it is only able to have one threshold causing a jagged line and the resulting AUC of the curves would not be accurate. We attempt to fix this for Laplacian and Laplacian of Gaussian by creating zero crossing manually. 

\subsection{Zero Crossing with Thresholdable Intensity}
We created our own zero crossing function that preserves intensity so that it can be thresholded at different values. This function detects if there is a sign change from the neighbouring pixels. If yes, there is a zero crossing and we assign the absolute value of the intensity of the pixel. Because we used absolute value of the intensity of the pixel, this will cause the ROC to be smoother since the thresholds can be continuous. This is illustrated in Listing \ref{customzero}.

\begin{lstlisting}[style=Matlab-editor, label={customzero}, caption={Customized zerocross function}]
function z_c_image = zero_cross_detection(image)
  [m, n] = size(image);
  z_c_image = zeros(m, n);
  for i = 1:m-1
      for j = 1:n-1
          if image(i,j) > 0
              if image(i+1,j) < 0 || image(i+1,j+1) < 0 || image(i,j+1) < 0
                  z_c_image(i,j) = abs(image(i,j));
              end
          elseif image(i,j) < 0
              if image(i+1,j) > 0 || image(i+1,j+1) > 0 || image(i,j+1) > 0
                  z_c_image(i,j) = abs(image(i,j));
              end
          end
      end
  end
end
\end{lstlisting}

\begin{figure}[H]
  \centering
  \begin{subfigure}[t]{0.40\textwidth}
    \includegraphics[width=\linewidth, height=6cm]{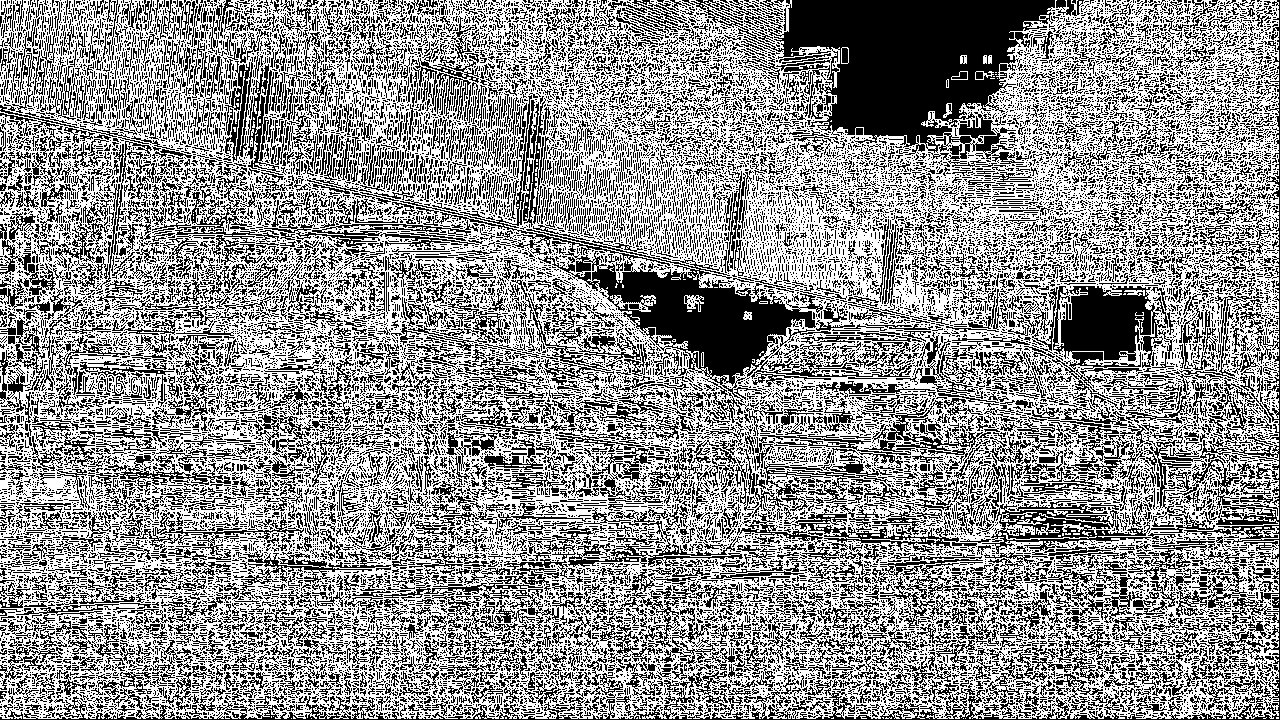}
    \caption{Laplacian}
  \end{subfigure}
  \begin{subfigure}[t]{0.40\textwidth}
    \includegraphics[width=\linewidth, height=6cm]{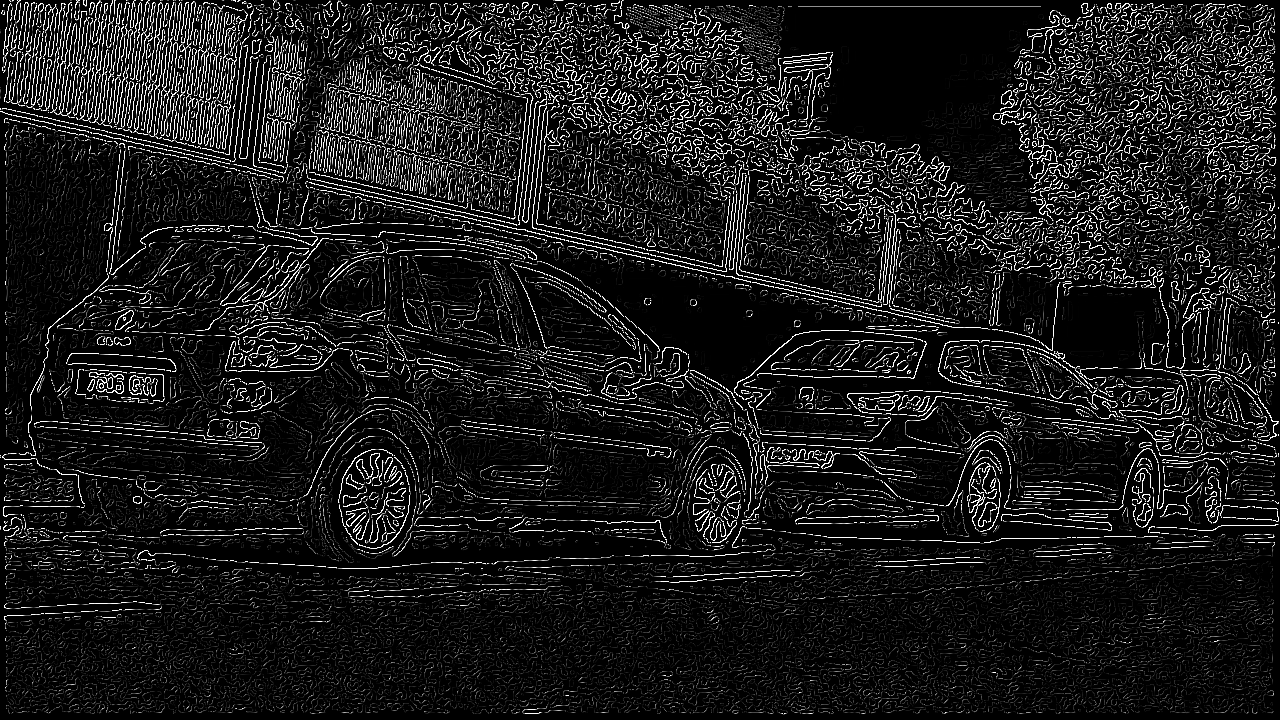}
    \caption{Laplacian of Gaussian}
  \end{subfigure}

  \caption{Laplacian and Laplacian of Gaussian after new zero crossing}
  \label{laplogzero}
\end{figure}

After creating this new zero crossing function we use this on RGB\_001 with Laplacian and Laplacian of Gaussian in Figure \ref{laplogzero} to show the effect. 
We observe that the image for Laplacian does not show good result. However, if we simply apply a threshold of 8 to the image then we get better result.

\begin{figure}[H]
  \centering
  \includegraphics[width=10cm, height=6cm]{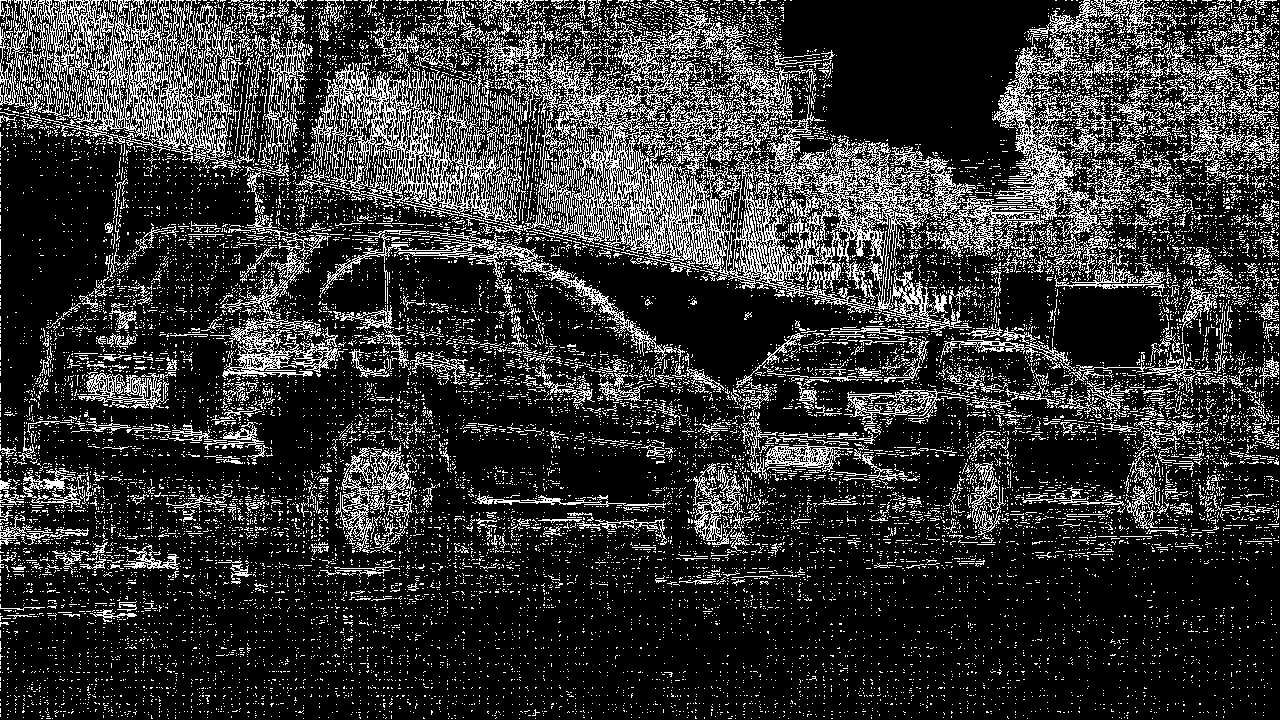}
  \caption{Laplacian after thresholding}
\end{figure}

\begin{figure}[H]
  \centering
  \includegraphics[width=10cm, height=6cm]{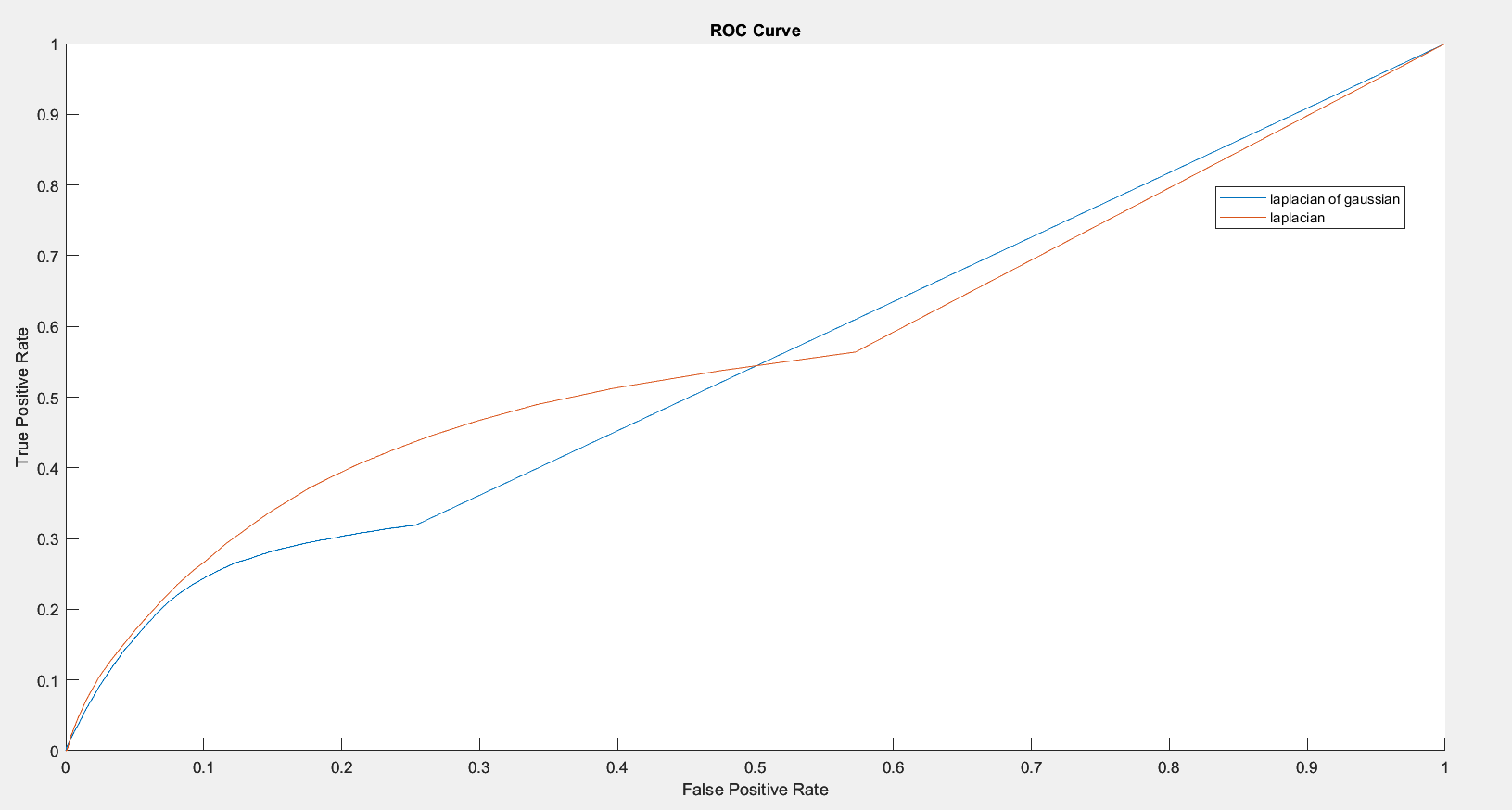}
  \caption{ROC for Laplacian and Laplacian of Gaussian after customized zero crossing}
  \label{roczero}
\end{figure}

We plotted the ROC curve and AUC \cite{roc} for the image RGB\_001, which can be seen in Figure \ref{roczero}. We observed that the curves were initially smooth but became jagged at certain points. This indicates the existence of a threshold where the Laplacian and Laplacian of Gaussian \cite{laplaciancite,log} images contains only 1, suggesting that their performance is only marginally better than random guessing. Table \ref{auczero} presents the AUC values for Laplacian and Laplacian of Gaussian, which remained relatively unchanged and slightly worse at 0.57 for Laplacian and 0.55 for Laplacian of Gaussian. Due to these findings, we need to devise a better strategy for converting the images convolved with Laplacian or Laplacian of Gaussian into thresholdable outputs.

\begin{table}[H]
  
  \caption{AUC for Laplacian and Laplacian of Gaussian after customized zero crossing}
  \centering
  \begin{tabular}{ |p{2cm}||p{2cm}|p{2cm}| }
      \hline
      \multicolumn{3}{|c|}{AUC} \\
      \hline
      Image & Laplacian & LoG  \\
      \hline
      RGB\_001 & 0.57 & 0.55 \\
      \hline
  \end{tabular}
  \label{auczero}
\end{table}

\subsection{Magnitude Thresholding for Laplacian and Laplacian of Gaussian}
Instead of using zero crossing to threshold at values, we could calculate the magnitude for Laplacian or Laplacian of Gaussian filtered image just as we did for Sobel, Roberts filter \cite{sobellap}. To do this, we will combine Laplacian with Sobel. We first convolve input image with Laplacian, then we convolve the output with Sobel filter to get gradients on both x and y-axis. After this we calculate the magnitude (Listing \ref{magnitude}) of gradients using absolute value. Listing \ref{lapmag} illustrates this.

\begin{lstlisting}[style=Matlab-editor, label={lapmag},caption={Magnitude of Laplacian}]
  Gx = conv2(laplacian_image, sobelX, 'same');
  Gy = conv2(laplacian_image, sobelY, 'same');
  G = magnitude(Gx, Gy);
\end{lstlisting}

\begin{figure}[H]
  \centering
  \begin{subfigure}[t]{0.40\textwidth}
    \includegraphics[width=\linewidth, height=6cm]{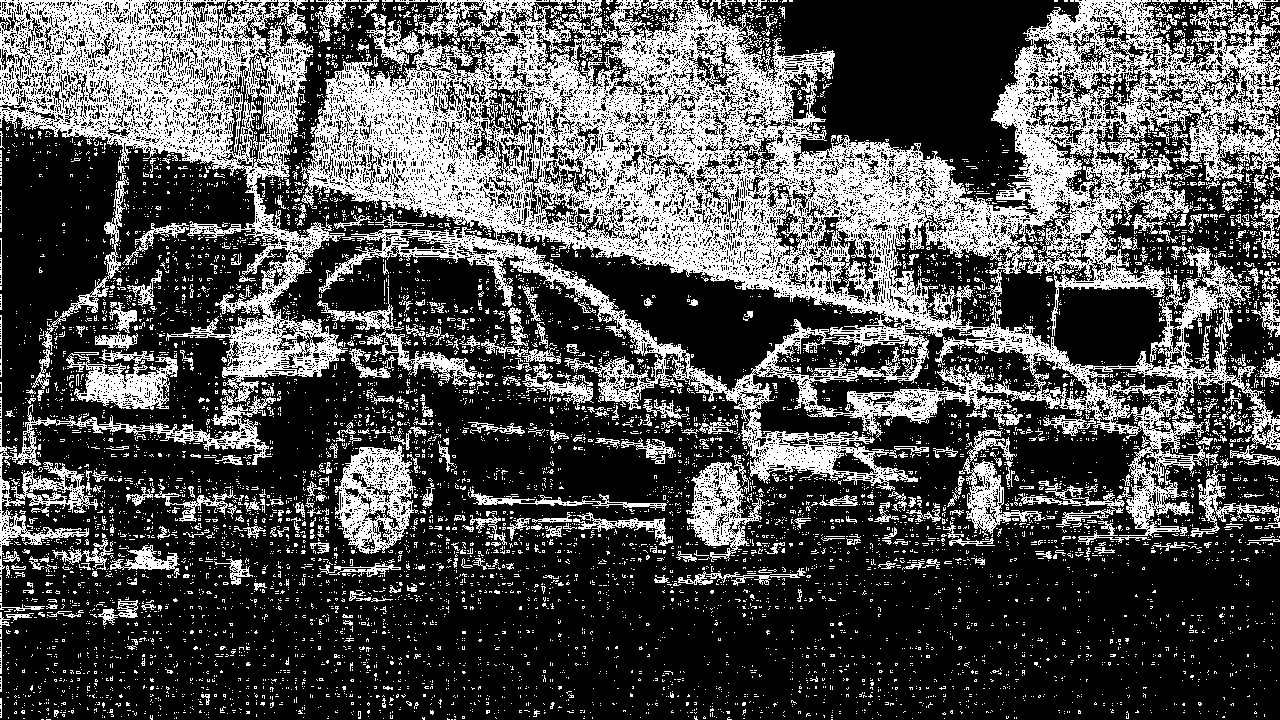}
    \caption{Laplacian with threshold of 40}
  \end{subfigure}
  \begin{subfigure}[t]{0.40\textwidth}
    \includegraphics[width=\linewidth, height=6cm]{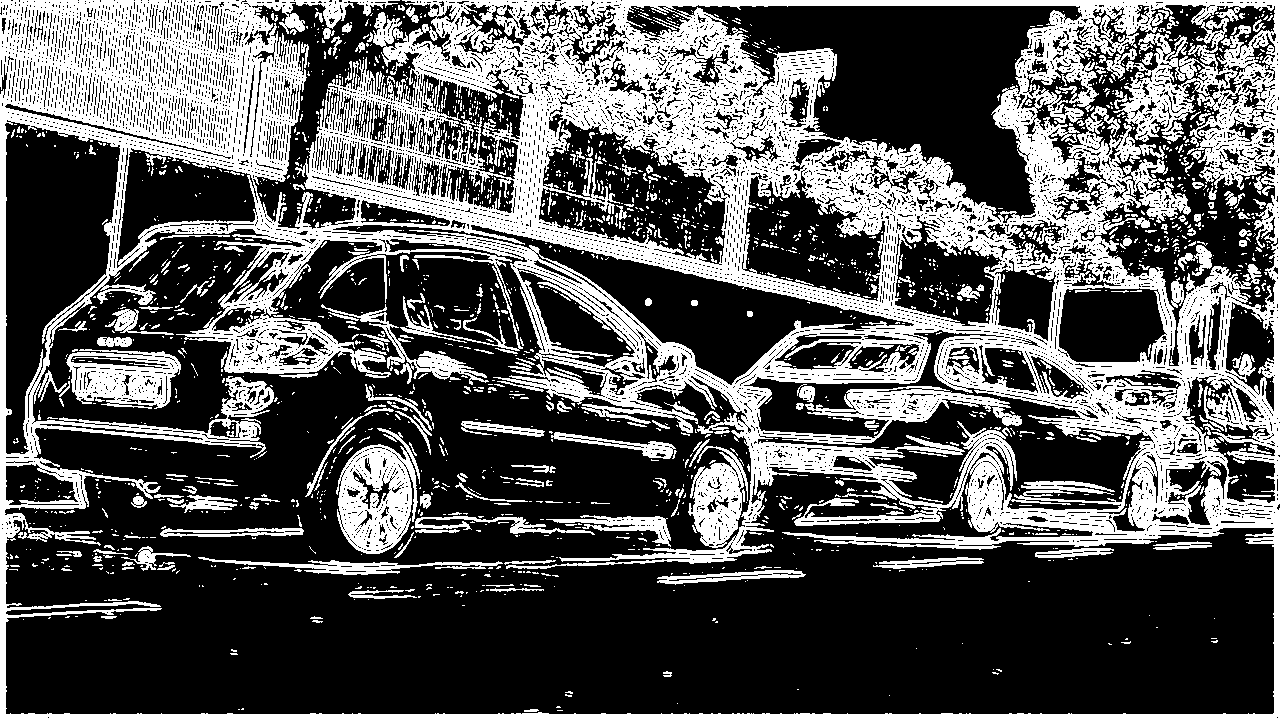}
    \caption{Laplacian of Gaussian with threshold of 8}
  \end{subfigure}
  \caption{Laplacian and Laplacian of Gaussian combined with Sobel to calculate magnitude}
  \label{magthresh}
\end{figure}

Figure \ref{magthresh} shows the results of customized Laplacian and Laplacian of Gaussian with thresholds 40 and 8 respectively on image RGB\_001.

\begin{figure}[H]
  \centering
  \includegraphics[width=10cm, height=6cm]{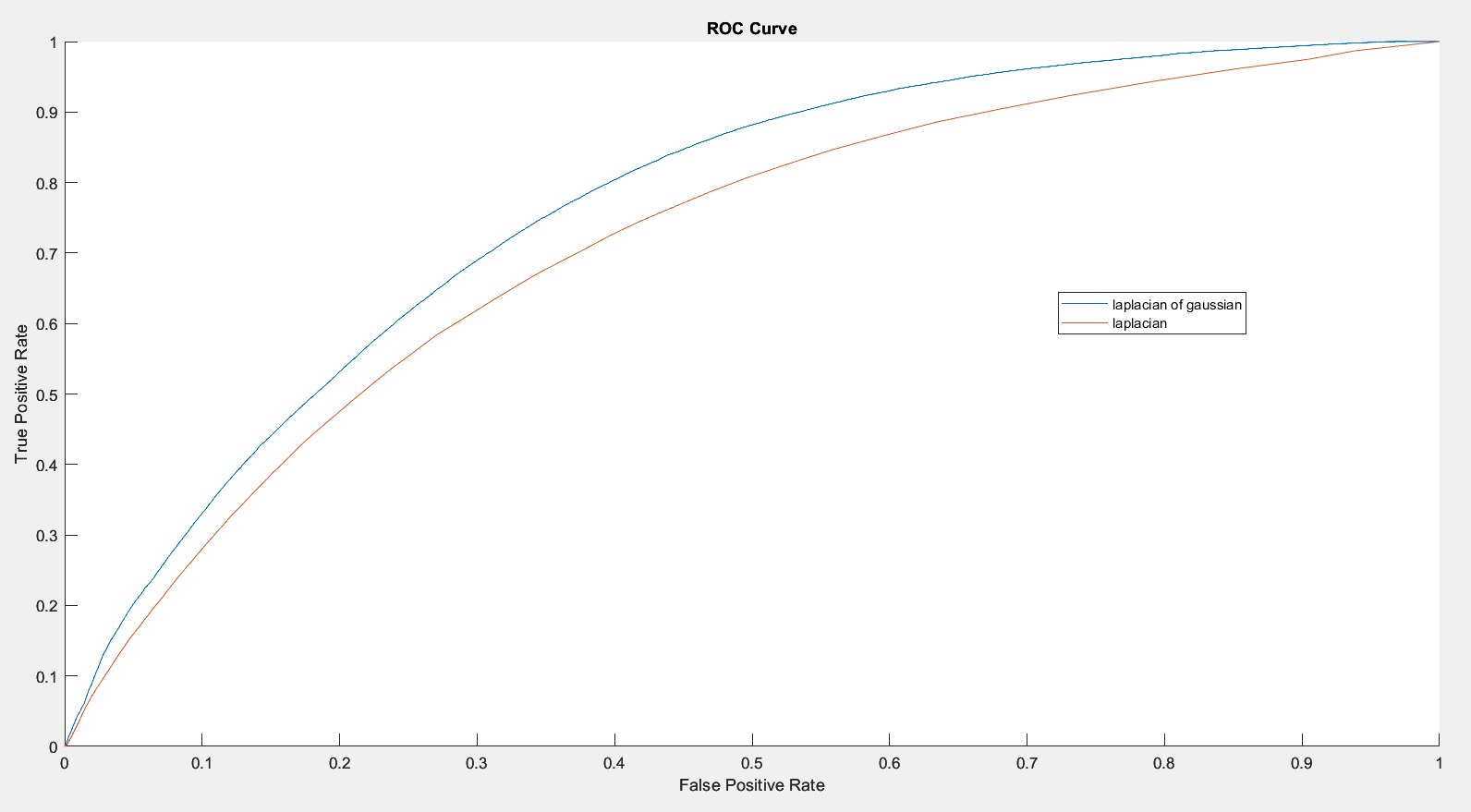}
  \caption{ROC after thresholding with magnitude}
  \label{rocmag}
\end{figure}

Figure \ref{rocmag} and Table \ref{aucmag} show the ROC and AUC of Laplacian and Laplacian of Gaussian combined with Sobel on image RGB\_001. We observe that it was a big improvement compared to zero crossing with AUC of 0.71 for Laplacian and 0.76 for Laplacian of Gaussian.

\begin{table}[H]
  
  \caption{AUC after thresholding with magnitude}
  \centering
  \begin{tabular}{ |p{2cm}||p{2cm}|p{2cm}| }
      \hline
      \multicolumn{3}{|c|}{AUC} \\
      \hline
      Image & Laplacian & LoG  \\
      \hline
      RGB\_001 & 0.71 & 0.76 \\
      \hline
  \end{tabular}
  \label{aucmag}
\end{table}

\subsection{Configuring Gaussian Kernel for Laplacian of Gaussian}
\begin{figure}[H]
  \centering
  \begin{subfigure}[t]{0.70\textwidth}
    \centering
    \includegraphics[width=10cm, height=6cm]{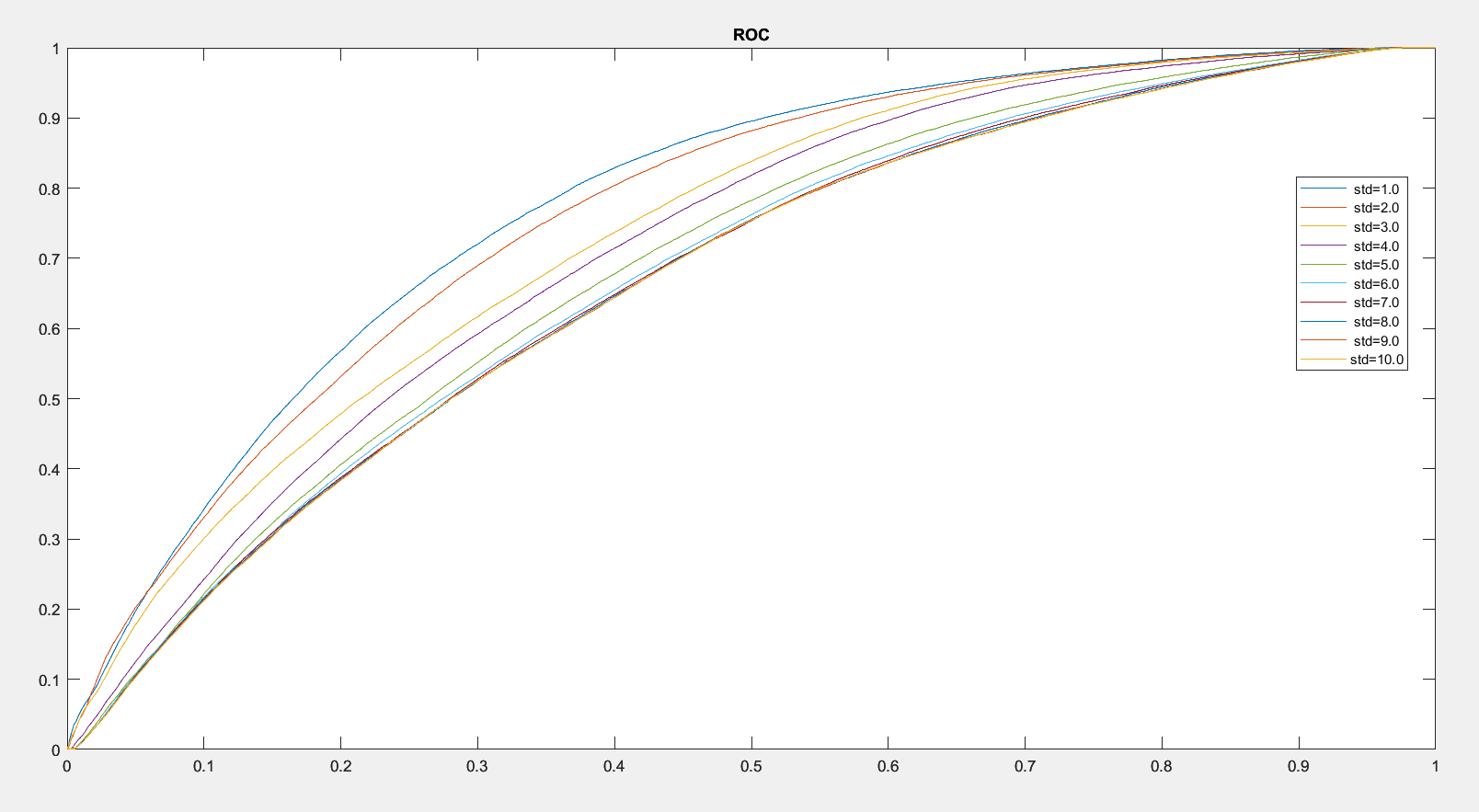}
    \caption{ROC curve for Laplacian of Gaussian with different standard deviations}
    \label{rocgaus}
  \end{subfigure}
  \begin{subfigure}[t]{0.20\textwidth}
    \includegraphics[width=\linewidth, height=4cm]{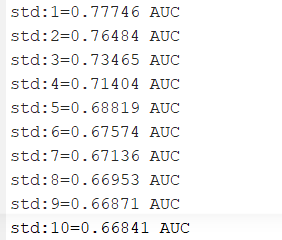}
    \caption{AUC for Laplacian of Gaussian with different standard deviations}
    \label{aucgaus}
  \end{subfigure}
  \caption{ROC and AUC of Laplacian of Gaussian with different standard deviations}

\end{figure}

\begin{figure}[H]
  \centering
  \begin{subfigure}[t]{0.70\textwidth}
    \includegraphics[width=\linewidth, height=6cm]{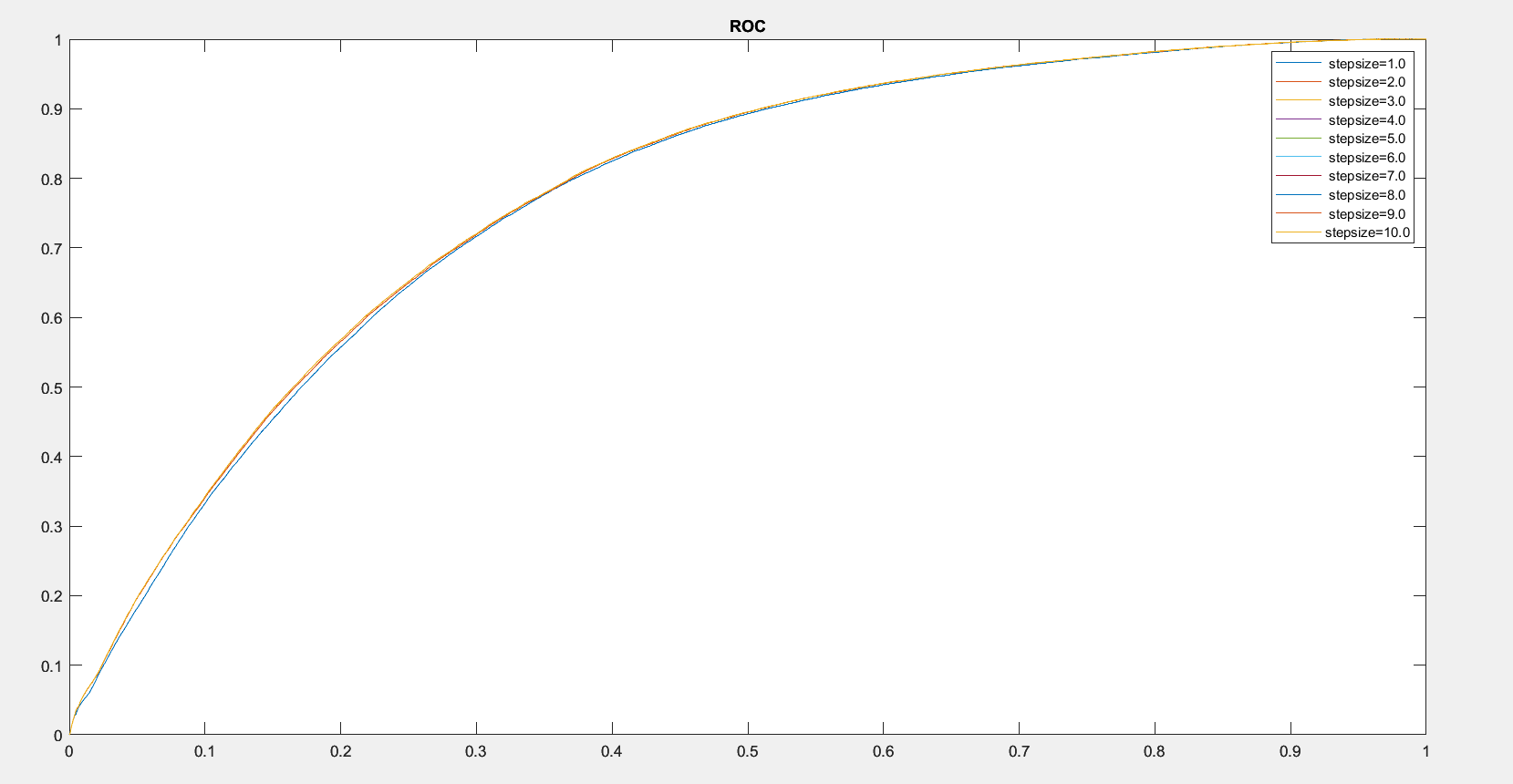}
    \caption{ROC curve for Laplacian of Gaussian with different stepsize}
    \label{rocstep}
  \end{subfigure}
  \begin{subfigure}[t]{0.20\textwidth}
    \includegraphics[width=\linewidth, height=4cm]{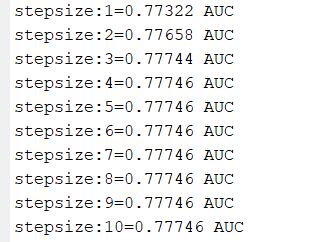}
    \caption{AUC of Laplacian of Gaussian with different stepsize}
    \label{aucstep}
  \end{subfigure}
  \caption{ROC and AUC of Laplacian of Gaussian with different stepsize}
\end{figure}

We can further configure the Laplacian of Gaussian \cite{log} by using different Gaussian kernels. Previously, we used a standard deviation of 2 and a mask size of 13x13 (stepsize = 6) for our Laplacian of Gaussian. To find the optimal configuration, we conducted an experiment varying both parameters. Figure \ref{rocgaus} displays the ROC curve of the Laplacian of Gaussian with different standard deviations. We discovered that the best value for the standard deviation is 1, resulting in an AUC of 0.78, as shown in Figure \ref{aucgaus}. After fixing this value, we experimented with different stepsizes. Figures \ref{rocstep} and \ref{aucstep} show the performance of Laplacian of Gaussian with different stepsize. We notice it starts performing best at stepsize 4.

\subsection{Canny Magnitude}
In addition to Laplacian \cite{laplaciancite} and Laplacian of Gaussian, we can manually configure the Canny \cite{canny} detector to obtain appropriate ROC and AUC values since the edge function only returns a binary image. After determining the optimal Gaussian kernel settings for ROC and AUC for Laplacian of Gaussian, with a standard deviation of 1 and a stepsize of 4, we reused the kernel for this task. Then we apply the Sobel filter, followed by non-maxima suppression. The last step involves hysteresis thresholding. To create the ROC curve \cite{roc}, we configure the hysteresis thresholding with both a low threshold and a high threshold. However, evaluating every possible combination of low and high thresholds would be impractical. Instead, we iterate through the range of low thresholds and set the high threshold to be double the value of the low threshold. Listing \ref{cannyroc} demonstrates the creation of the ROC curve using this method. We use a vector \texttt{lowT} to store the low thresholds, which are evenly spaced using the \texttt{linspace} function. It is important to note that \texttt{maxmag} represents the maximum magnitude of the gradients in the smoothed image, allowing the low thresholds to range from \texttt{0} to \texttt{maxmag/2}. For each low threshold, we set the high threshold \texttt{valueH} to be double the low threshold \texttt{valueL}. Then we detect the edges using the Canny detection algorithm created from scratch and compute the true positive rate (TPR) and false positive rate (FPR). These TPR and FPR values are stored in vectors for further analysis. The complete Canny algorithm from scratch can be found in Appendix \ref{cannyalgo}.

\begin{lstlisting}[style=Matlab-editor, label={cannyroc}, caption={Code snippet of creating ROC curve of canny algorithm}]
  numtrials =80;
  TPR = zeros(numtrials,1);
  FPR = zeros(numtrials,1);
  % low threshold we scatter all possible low threshold as linear space
  lowT = linspace(0,maxmag/2,numtrials);
  for x = 1:numtrials
      valueL = lowT(x);
      valueH = valueL *2;
      if valueH < maxmag
         
          predicted = cannyEdgeDetection(image, gaussian_filter, valueL, valueH);
          predicted = predicted(:);
          # compute true positive rate and false positive rate 
          TPR(x)= sum(gt_arr == 1 & predicted == gt_arr) ./ sum(gt_arr ==1);
          FPR(x) = sum(gt_arr ==0 & predicted ~= gt_arr) ./ sum(gt_arr == 0);
      end
  end
  plot(FPR, TPR);
\end{lstlisting}

\begin{figure}[H]
  \centering
  \includegraphics[width=8cm, height = 4cm]{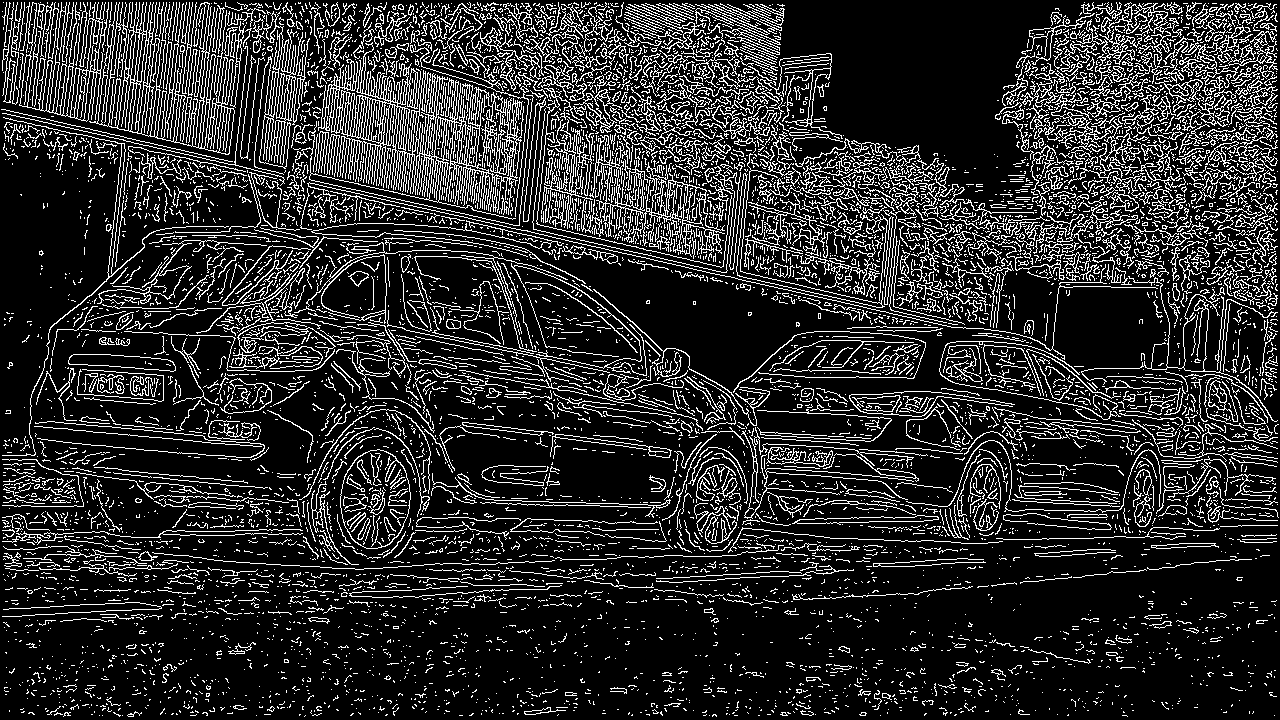}
  \caption{Detected edges using Canny}
  \label{cannyedge}
\end{figure}

Figure \ref{cannyedge} shows the detected edges using Canny detection \cite{canny} with low and high thresholds of 10 and 20 respectively.

\begin{figure}[H]
 
  \centering
  \includegraphics[width = 10cm, height = 5cm]{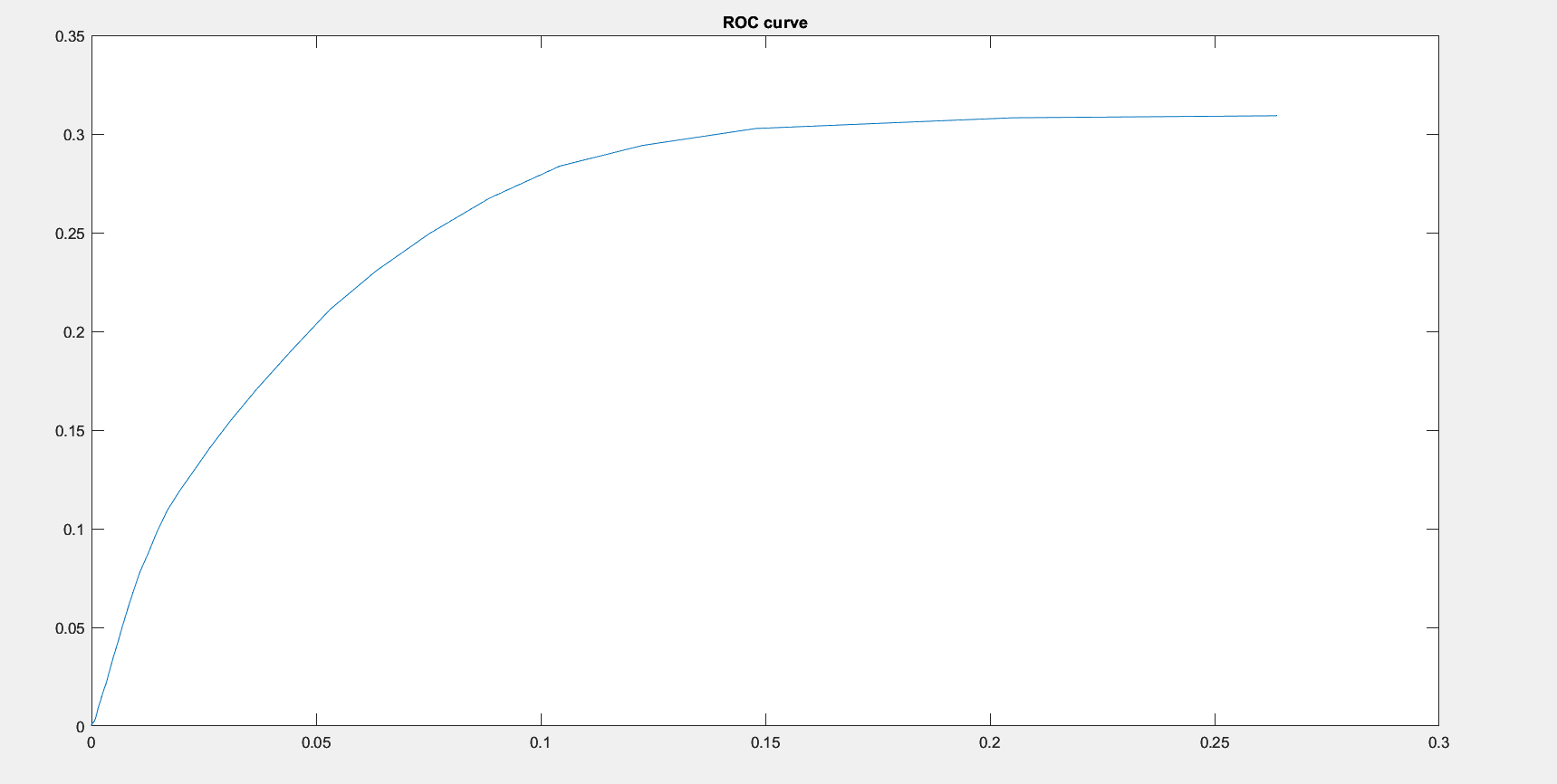}
  \caption{ROC curve for Canny algorithm}
  \label{rocnonmaxima}
\end{figure}

There is a fundamental problem here. We plotted the ROC and noticed that the maximum TPR was very low, approximately 0.30. Figure \ref{rocnonmaxima} illustrates this. 

\begin{figure}[H]
  \centering
  \includegraphics[width=8cm, height = 4cm]{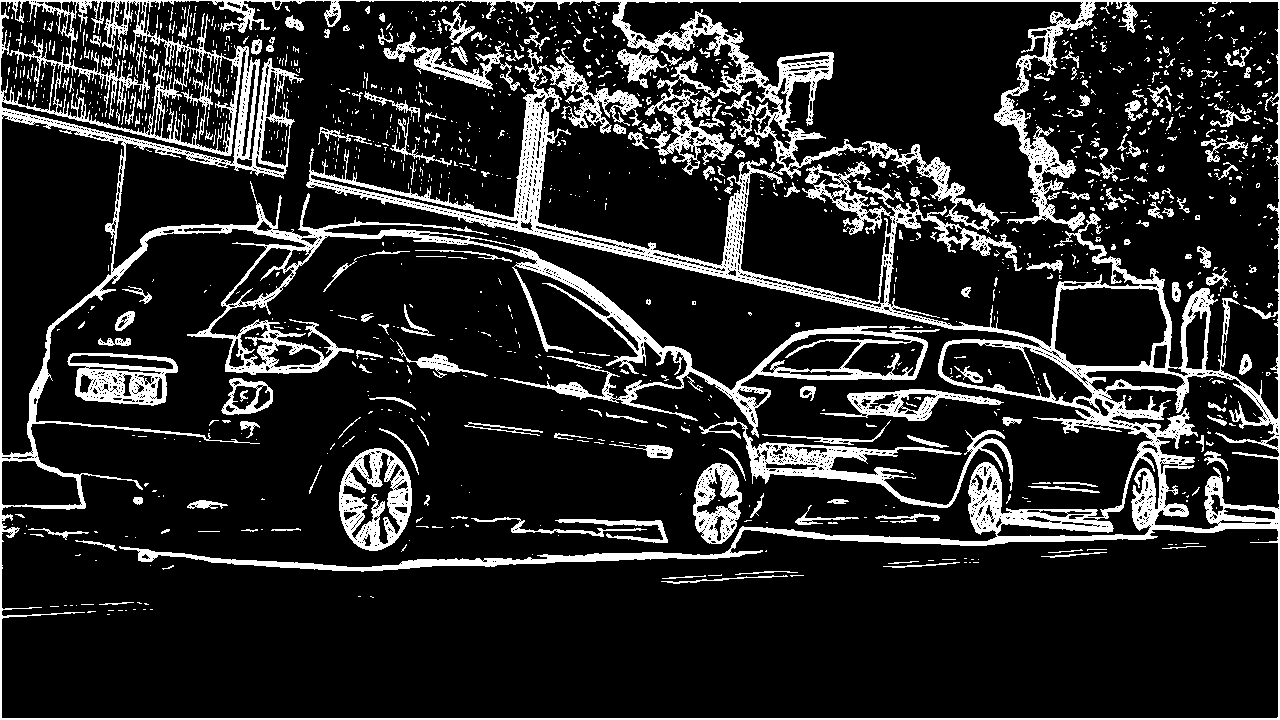}
  \caption{Canny edge detection without non-maximum detection}
  \label{cannyedgerevised}
\end{figure}

It is evident that the performance of the edge detector is significantly worse compared to other edge detectors. Upon investigation, we discovered that the culprit behind this poor performance is the non-maximum suppression step. Non-maximum suppression aims to thin the edges by removing edge points with magnitudes lower than the local maximum. However, this approach can introduce a shift in the detected edges relative to the ground truth image, as the local maximum of the edge may differ from the ground truth edge. Consequently, the true positive rate (TPR) is affected. To further explore this issue, we conducted an experiment where we removed the non-maximum suppression from the Canny detector \cite{canny} and plotted the ROC curve \cite{roc}. This is why we created Canny algorithm from scratch, enabling us to remove the non-maximum suppression step. Figure \ref{cannyedgerevised} illustrates the detected edges of Canny without non-maximum suppression.

\begin{figure}[H]
  \centering

  \begin{subfigure}[t]{0.70\textwidth}
    \includegraphics[width=\linewidth, height=6cm]{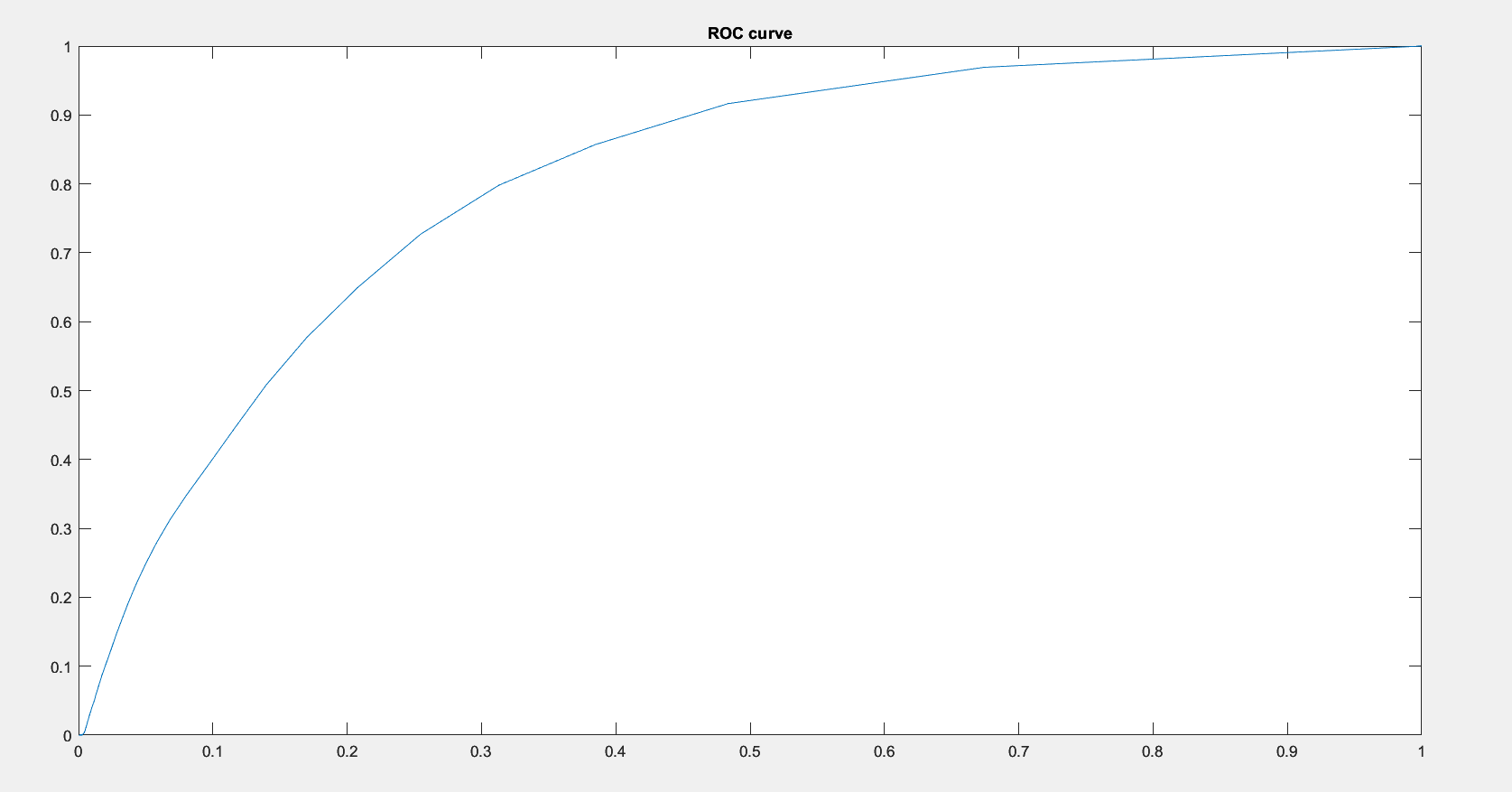}
    \caption{ROC curve for Canny without non-maximum suppression}
    \label{roccanny}
  \end{subfigure}
  \begin{subfigure}[t]{0.20\textwidth}
    \includegraphics[width=\linewidth, height=1cm]{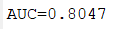}
    \caption{AUC for Canny without non-maximum suppression}
    \label{auccanny}
  \end{subfigure}
  \caption{ROC and AUC of Canny without non-maximum suppression}
\end{figure}

Figures \ref{roccanny} and \ref{auccanny} show the ROC and AUC of Canny without non-maximum suppression. We immediately see that the ROC curve was much better than previous one and AUC increased to 0.80. We therefore conclude that non-maximum suppression did cause low performance for Canny algorithm.

\section{Final Results}
\begin{figure}[H]
  \centering
  \begin{subfigure}[t]{0.40\textwidth}
    \includegraphics[width=\linewidth, height=6cm]{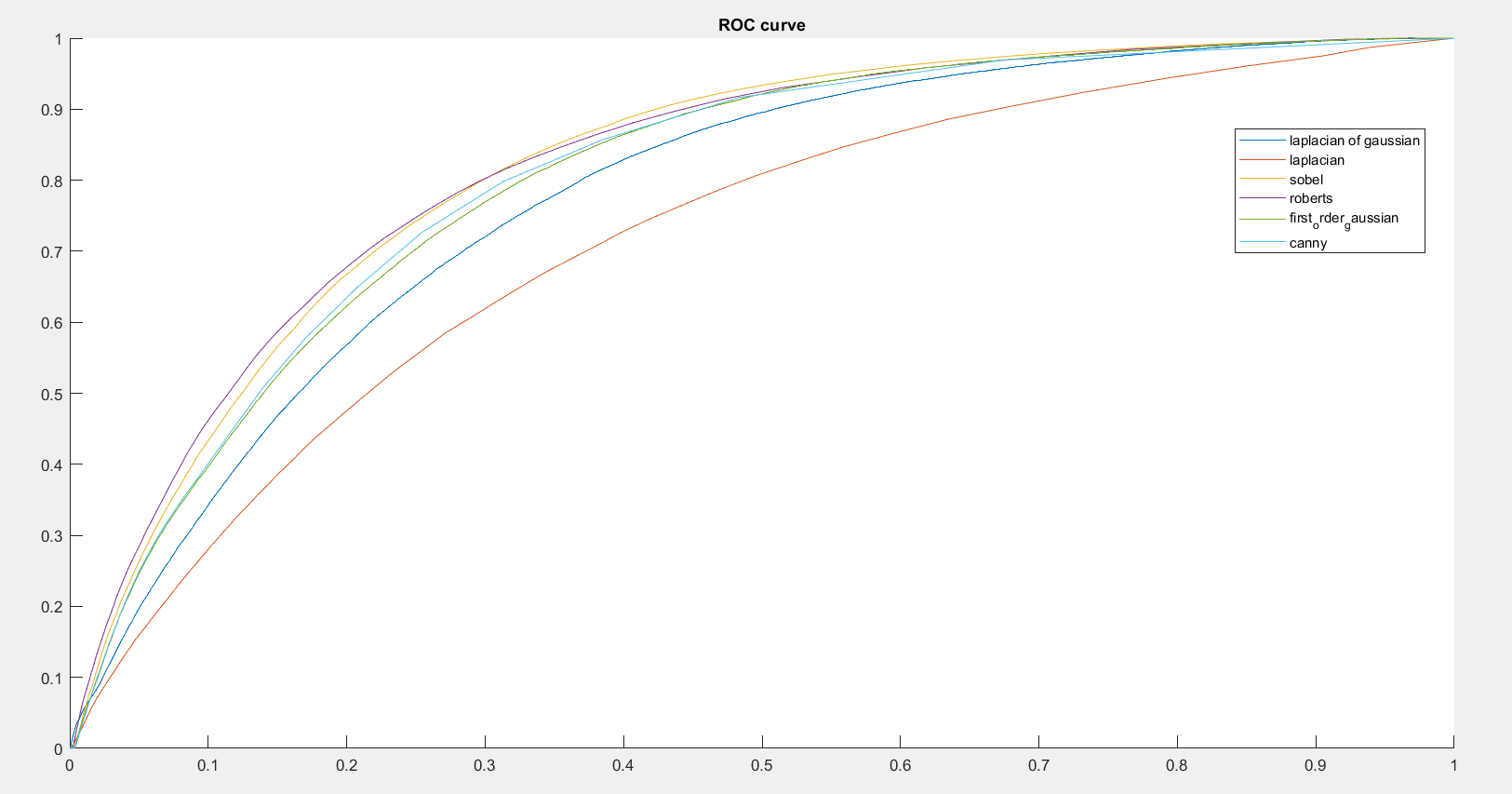}
    \caption{RGB\_001}
  \end{subfigure}
  \begin{subfigure}[t]{0.40\textwidth}
    \includegraphics[width=\linewidth, height=6cm]{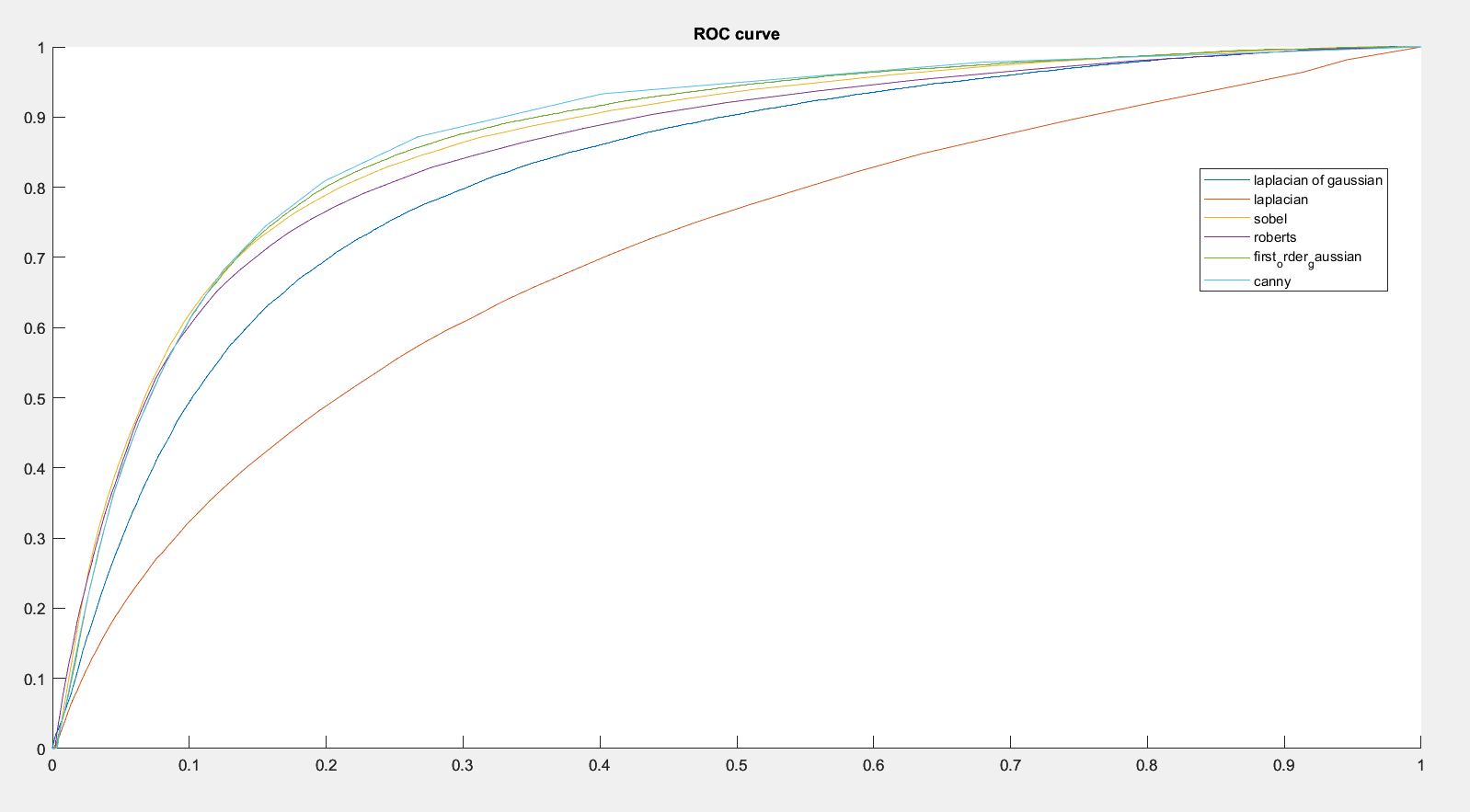}
    \caption{RGB\_002}
  \end{subfigure}
  \begin{subfigure}[t]{0.40\textwidth}
    \includegraphics[width=\linewidth, height=6cm]{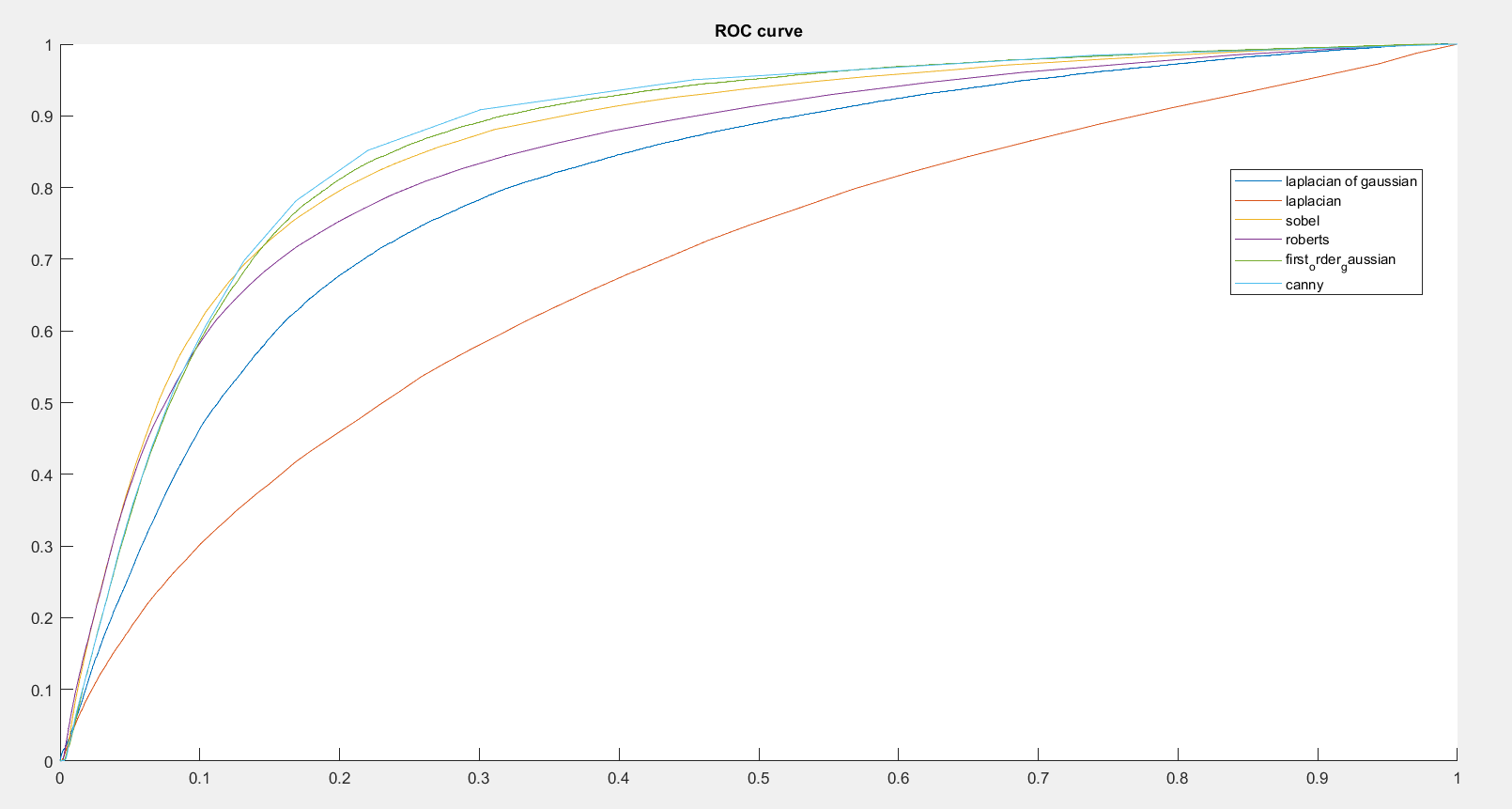}
    \caption{RGB\_003}
  \end{subfigure}
  \caption{ROC curves of configured algorithms}
  \label{rocrevised}
\end{figure}

After configuring Laplacian, Laplacian of Gaussian and Canny edge detectors \cite{laplaciancite,log,canny}, we now can construct the ROC and AUC \cite{roc} of all the algorithms for all 3 images from BIPED \cite{soria2020dense} dataset. Figure \ref{rocrevised} and Table \ref{aucrevised} show the ROC and AUC of the three images respectively. We observe that for RGB\_001 Roberts and Sobel filter performed best with AUC of 0.82. For RGB\_002 Canny had impressive results with AUC of 0.87. Lastly, First-order Gaussian \cite{firstorder} and Canny \cite{canny} performed best for RGB\_003. Both had 0.87 AUC. We notice that Laplacian filter \cite{laplaciancite} performed worst for all images. The AUC increased when using Laplacian of Gaussian.

\begin{table}[H]
  \caption{AUC of configured edge detectors}
  \begin{tabular}{ |p{2cm}||p{2cm}|p{2cm}|p{2cm}||p{2cm}|p{2cm}|p{2cm}|  }
      \hline
      \multicolumn{7}{|c|}{AUC} \\
      \hline
      Image & Roberts & Sobel & FoG & Laplacian & LoG & Canny \\
      \hline
      RGB\_001 & 0.82 & 0.82 & 0.80 & 0.71 & 0.78 & 0.80 \\
      RGB\_002 & 0.85 & 0.86 & 0.86 & 0.70 & 0.82 & 0.87 \\
      RGB\_003 & 0.84 & 0.86 & 0.87 & 0.69 & 0.80 & 0.87\\
      \hline
  \end{tabular}
  \label{aucrevised}
\end{table}

\section{Discussion}
The customization of certain edge detectors allowed us to improve their ROC and AUC performance \cite{roc}. Initially, we experimented with zero crossing for Laplacian and Laplacian of Gaussian \cite{laplaciancite,log}, but we found that it alone did not result in a smooth ROC curve. To enhance the performance, we combined Laplacian and Laplacian of Gaussian with first-order derivative filters, specifically the Sobel filter \cite{sobellap}, which enabled thresholding based on magnitude. This integration yielded a smoother curve and improved results.

In the case of Laplacian of Gaussian \cite{log}, we further conducted an experiment to identify the optimal Gaussian kernel by varying the standard deviation and stepsize. It is worth noting that the optimal configuration may vary depending on the specific circumstances.

For the Canny edge detector \cite{canny}, we addressed the challenge of configuring hysteresis thresholding. Since evaluating every combination of low and high thresholds is impractical, we set the high threshold to double the value of the low threshold. However, we discovered that the performance was hindered by the non-maximum suppression step, resulting in a true positive rate (TPR) of approximately 0.30 with respect to the ground truth image. To improve the results, we removed the non-maximum suppression step, leading to a significant performance enhancement.

As a direction for future research, it would be valuable to investigate the optimal relationship between the low and high thresholds in hysteresis thresholding. Exploring this aspect can provide further insights into improving the performance of the Canny edge detector.

\section{Conclusion}
In conclusion, our study demonstrated that customization techniques can significantly improve the ROC and AUC performance of certain edge detectors. By combining Laplacian and Laplacian of Gaussian with first-order derivative filters, we achieved smoother ROC curves and improved results \cite{sobellap}. Additionally, our experimentation with different Gaussian kernels for Laplacian of Gaussian \cite{log} highlighted the importance of finding the optimal configuration specific to each situation. We also found that removing the non-maximum suppression step in the Canny edge detector substantially enhanced its performance. Overall, our findings contribute to the advancement of edge detection algorithms and pave the way for more effective image analysis and computer vision applications

\bibliography{library}

\begin{thebibliography}{10}

\bibitem{firstorder}
Jassim Abdul-Jabbar and Abdulhamed M.~Jasim.
\newblock Design and multiplierless implementations of ecg-based 1st order
  gaussian derivative wavelet filter with lattice structures.
\newblock {\em Journal of University of Anbar for Pure Science}, 6, 08 2012.

\bibitem{roc}
Andrew~P. Bradley.
\newblock The use of the area under the {ROC} curve in the evaluation of
  machine learning algorithms.
\newblock {\em Pattern Recognition}, 30(7):1145--1159, 1997.

\bibitem{canny}
John Canny.
\newblock A computational approach to edge detection.
\newblock {\em IEEE Transactions on Pattern Analysis and Machine Intelligence},
  PAMI-8(6):679--698, 1986.

\bibitem{laplaciancite}
Zhenlong Du and Xiaoli Li.
\newblock Laplacian filtering effect on digital image tuning via the decomposed
  eigen-filter.
\newblock 78, 09 2019.

\bibitem{MATLAB}
The~MathWorks Inc.
\newblock Matlab version: 9.13.0 (r2022b), 2022.

\bibitem{sobel}
N.~Kanopoulos, N.~Vasanthavada, and R.L. Baker.
\newblock Design of an image edge detection filter using the {Sobel} operator.
\newblock {\em IEEE Journal of Solid-State Circuits}, 23(2):358--367, April
  1988.
\newblock Conference Name: IEEE Journal of Solid-State Circuits.

\bibitem{sobellap}
{Mewar University, Chittorgarh, India}, Suneet Gupta, Rabins Porwal, and
  {International College of Engineering, India}.
\newblock {COMBINING} {LAPLACIAN} {AND} {SOBEL} {GRADIENT} {FOR} {GREATER}
  {SHARPENING}.
\newblock {\em ICTACT Journal on Image and Video Processing},
  06(04):1239--1243, May 2016.

\bibitem{roberts}
Malik Sagheer, Chun He, Nicola Nobile, and Ching Suen.
\newblock Holistic urdu handwritten word recognition using support vector
  machine.
\newblock pages 1900--1903, 08 2010.

\bibitem{soria2020dense}
Xavier Soria, Edgar Riba, and Angel~D. Sappa.
\newblock Dense extreme inception network: Towards a robust cnn model for edge
  detection, 2020.

\bibitem{log}
G.E. Sotak and K.L. Boyer.
\newblock The laplacian-of-gaussian kernel: A formal analysis and design
  procedure for fast, accurate convolution and full-frame output.
\newblock {\em Computer Vision, Graphics, and Image Processing},
  48(2):147--189, 1989.

\end{thebibliography}
\bibliographystyle{plain}

\appendix
\section*{Appendix}
\section{Canny algorithm from scratch}
\label{cannyalgo}
\begin{lstlisting}[style=Matlab-editor, label={cannyfunction}, caption={Code snipped of Canny algorithm}]
  function [edges, maxmag] = cannyEdgeDetection(image, gaussian_kernel, lowThreshold, highThreshold)
  smoothedImage = conv2(image, gaussian_kernel, 'same');
  load filters;
  Gx = conv2(smoothedImage, sobelX, 'same');
  Gy = conv2(smoothedImage, sobelY, 'same');
  gradientMagnitude = magnitude(Gx, Gy);
  maxmag = max(gradientMagnitude(:));
  gradientDirection = atan2(Gy, Gx);
  suppressedImage = nonMaxSuppression(gradientMagnitude, gradientDirection);
  edges = hysteresisThresholding(suppressedImage, lowThreshold, highThreshold);
end

function suppressedMagnitude = nonMaxSuppression(gradientMagnitude, gradientDirection)
  [rows, cols] = size(gradientMagnitude);
  suppressedMagnitude = zeros(rows, cols);
  

  gradientDirection = gradientDirection * 180 / pi;
  gradientDirection(gradientDirection < 0) = gradientDirection(gradientDirection < 0) + 180;
  for i = 2:rows-1
      for j = 2:cols-1
          direction = gradientDirection(i, j);
          
          if (0 <= direction && direction < 22.5) || (157.5 <= direction && direction < 180)
              q = gradientMagnitude(i, j+1);
              r = gradientMagnitude(i, j-1);
          elseif 22.5 <= direction && direction < 67.5
              q = gradientMagnitude(i+1, j-1);
              r = gradientMagnitude(i-1, j+1);
          elseif 67.5 <= direction && direction < 112.5
              q = gradientMagnitude(i+1, j);
              r = gradientMagnitude(i-1, j);
          elseif 112.5 <= direction && direction < 157.5
              q = gradientMagnitude(i-1, j-1);
              r = gradientMagnitude(i+1, j+1);
          end
          
          if gradientMagnitude(i, j) >= q && gradientMagnitude(i, j) >= r
              suppressedMagnitude(i, j) = gradientMagnitude(i, j);
          end
      end
  end
end

function edges = hysteresisThresholding(suppressedMagnitude, lowThreshold, highThreshold)
  [rows, cols] = size(suppressedMagnitude);
  edges = zeros(rows, cols);
  strongEdges = suppressedMagnitude > highThreshold;
  weakEdges = suppressedMagnitude > lowThreshold;
  strongEdges = bwmorph(strongEdges, 'bridge');
  edges = strongEdges & weakEdges;
end
\end{lstlisting}

\end{document}